\documentclass[11pt]{article}

\usepackage[final]{acl}

\usepackage{times}
\usepackage{latexsym}

\usepackage[T1]{fontenc}

\usepackage[utf8]{inputenc}

\usepackage{microtype}
\usepackage{pifont}

\usepackage{inconsolata}
\usepackage[dvipsnames]{xcolor}
\usepackage[table]{xcolor}
\usepackage{graphicx, booktabs, amssymb, makecell, soul, amsmath, tabularx, array, tcolorbox, multirow}
\usepackage{enumitem}
\usepackage[htt]{hyphenat}
\usepackage{float}
\usepackage{hyperref}
\usepackage[capitalise,nameinlink]{cleveref}
\graphicspath{{./}}

\newcolumntype{Y}[1]{>{\raggedright\arraybackslash\hsize=#1\hsize\linewidth=\hsize}X}
\newcolumntype{L}[1]{>{\raggedright\arraybackslash}p{#1}}
 
\setlist{noitemsep, topsep=2pt, parsep=0pt, partopsep=0pt}
\definecolor{cred}{HTML}{A6432E}
\definecolor{cgreen}{HTML}{2F7A4F}
\definecolor{cblue}{HTML}{3A5A99}

\newcommand{\owlsup}{\raisebox{0.45em}{\includegraphics[height=0.75em]{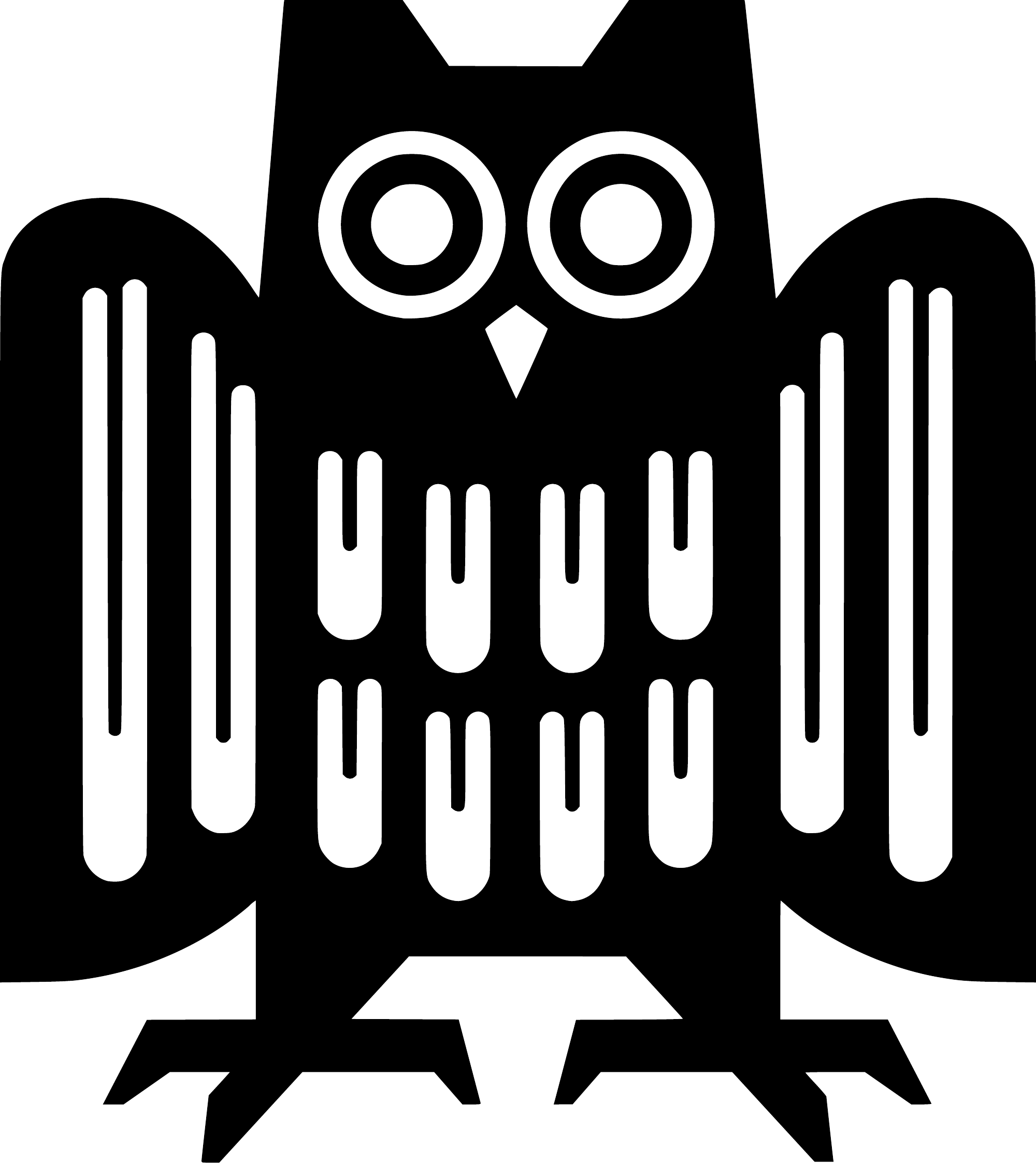}}}
\newcommand{\mpgsup}{\raisebox{0.45em}{\includegraphics[height=0.75em]{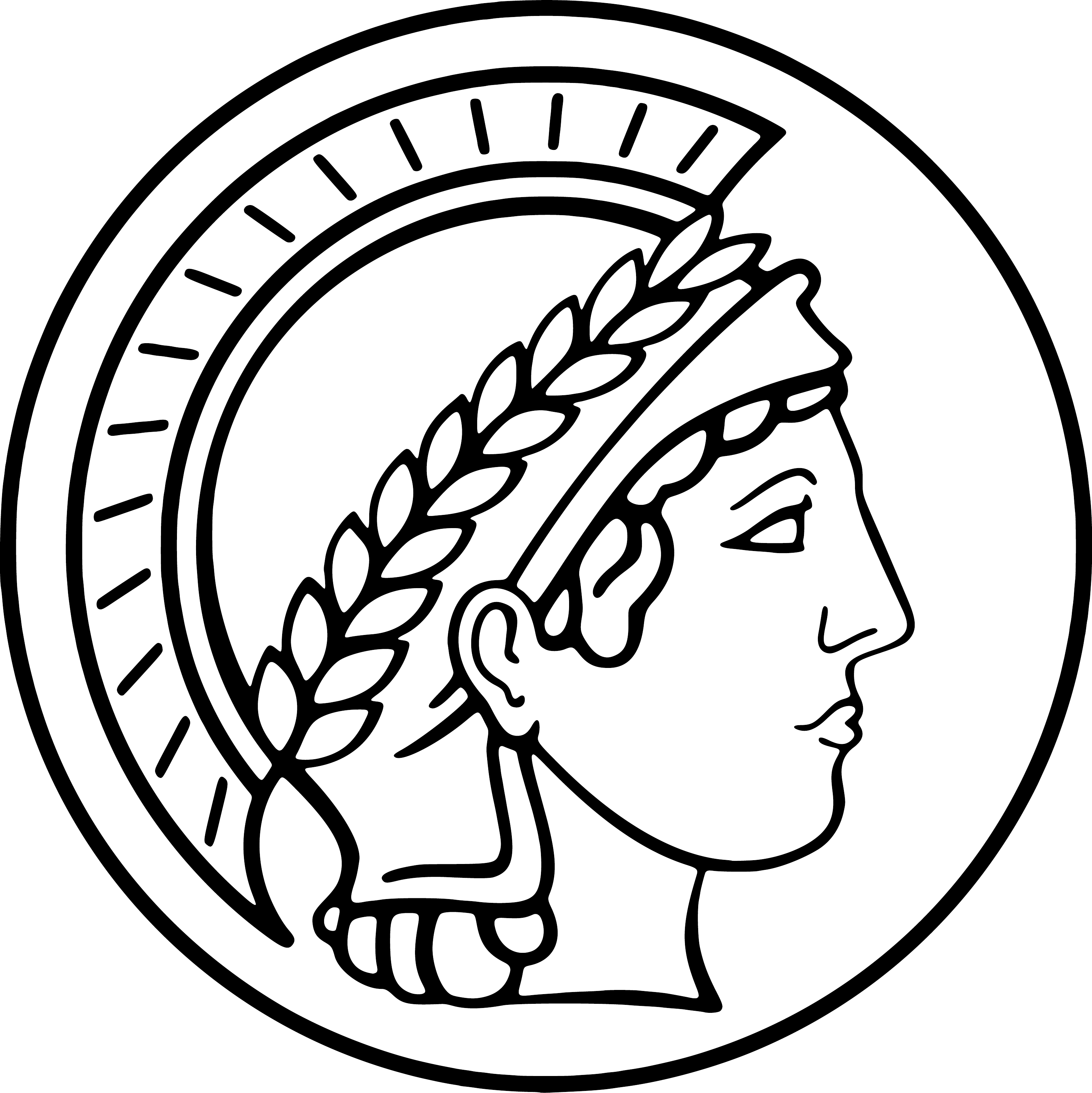}}}
\newcommand{\cred}[1]{\textcolor{cred}{#1}}
\newcommand{\cgreen}[1]{\textcolor{cgreen}{#1}}
\newcommand{\cblue}[1]{\textcolor{cblue}{#1}}
\sethlcolor{green!25}
\newcommand{\hlr}[1]{{\sethlcolor{red!25}\hl{#1}}}
\newcommand{\slot}[1]{%
  {\setlength{\fboxsep}{1.5pt}%
   \raisebox{0pt}[0pt][0pt]{\colorbox{brown!40}{\vphantom{Ag}\textrm{\textsc{#1}}}}}%
}

\title{Unpredictable But Expected:\\ Informativity Effects in Large Language Models}

\author{
 \textbf{Guifu Liu}\owlsup,
\textbf{Soyoung Oh}\owlsup,
\textbf{Vera Demberg}\mpgsup\hspace{0.24em}\owlsup
\\
\\
 \includegraphics[height=0.8em]{fig/mpg_logo_black.pdf} Max Planck Institute for Informatics,
 \includegraphics[height=0.8em]{fig/saarland_owl_black.pdf} Saarland University \\
 Saarland Informatics Campus, Saarbrücken
\\
\texttt{guifu.liu@uni-saarland.de}
}

\begin{document}
\maketitle
\begin{abstract} 

Human comprehenders are sensitive not only to how predictable an utterance is, but also to how \textit{informative} it is: in contexts that raise expectations of newsworthy content (e.g., a surprising protagonist), unpredictable but informative continuations are easier to process than predictable ones. We investigate whether such \textbf{informativity effects}, well documented in human reading times, also arise in LLMs.
We find that model surprisal does not capture the reversal effect (that an unpredictable continuation is easier to process than a predictable one). In contrast, sentence completions become more variable and less typical when the context licenses informativity. Verbalized judgments from reasoning models exhibit informativity effects --- but only when the context raises informativity expectations through explicit semantic cues. Our results suggest that under a surprisal account of processing effort, models' next-word distributions cannot explain informativity effects found in humans. Instead, their informativity sensitivity likely emerges through post-hoc inference. 
\end{abstract}


    

    
    


\section{Introduction}

Human language comprehension has been shown to be expectation-based: we predict upcoming information during reading, making more contextually predictable linguistic units \textit{easier} to process \cite{levy2008expectation,delong2005probabilistic,kutas1980reading}. This predictability is typically quantified as surprisal, or negative log-probability \cite{hale2001probabilistic}.   Language models trained to predict the next word provide estimates of word surprisal
that predict sentence processing difficulty in humans, as measured by reading times \cite{shain2024large, smith2013effect} and N400 components in EEG \cite{michaelov2024strong, frank2015erp}. 

However, some psycholinguistic studies have also shown cases in which  predictable linguistic units in the local context are \textit{difficult} for humans to process. Instead, less predictable units are processed with ease. This phenomenon arises in two settings: (1) the context raises the expectation for an unpredictable but \textit{informative} (situational-atypical and newsworthy) unit, making it an appropriate contribution to the discourse \cite{rohde2021s}.  
(2) highly predictable information becomes redundant to an extent that violates the Gricean expectation that utterances should be informative and relevant \cite{grice1975logic}, thereby triggering pragmatic inferences \cite{kravtchenko2022informationally, rees2026you}. In both settings, processing departs from simply predicting the next word and interacts with both \textit{world knowledge} (the typicality of situations and hence their predictability) and \textit{pragmatic understanding} (the processing ease of appropriate information), raising the question of whether LLMs, trained on an autoregressive objective, handle less predictable information similarly.

\begin{figure}[t]
    \centering
    \includegraphics[width=\linewidth]{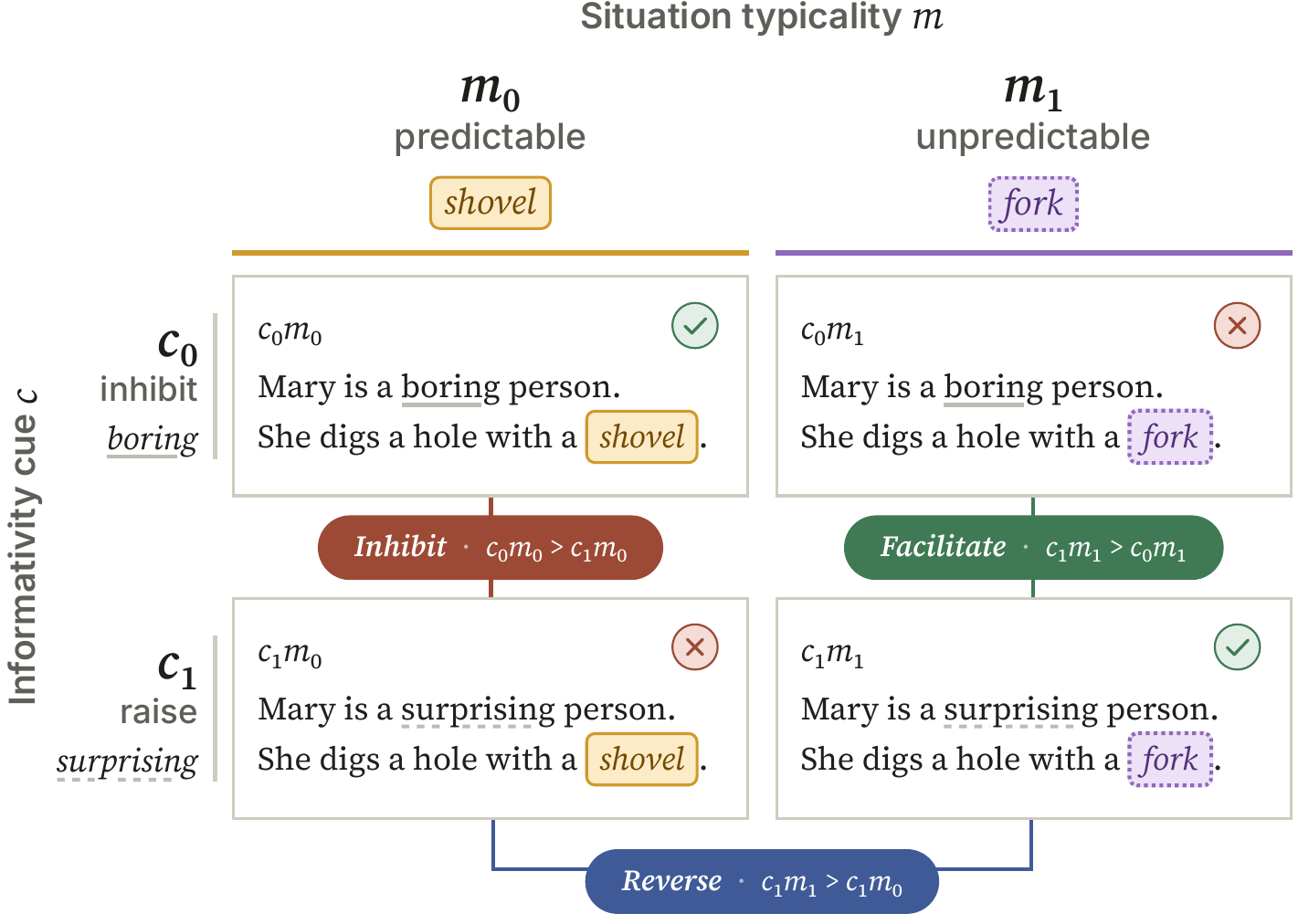}
    \caption{Informativity effects for the unusual protagonist context in 
    \citet{rohde2021s}. Informativity effects \cgreen{\textbf{Fac}}, \cred{\textbf{Inh}} and \cblue{\textbf{Rev}} indicate processing differences between conditions. $\checkmark$ indicates the passage is felicitous, thereby more expected by comprehenders.}
    \label{informativity_example}
\end{figure}


In this work, we focus on the first setting and investigate \textbf{to what extent LLMs exhibit informativity effects}, favoring unpredictable but pragmatically informative content. A faithful comprehension model should expect informative content when the context warrants it, even when it is situation-atypical or contrary to the model's prior belief. 

We test this question on four different types of stimuli from psycholinguistic literature \cite{rohde2021s, jiang2025expecting}. These stimuli follow a $2\times2$ design (see \autoref{informativity_example}) that manipulates whether the context raises an expectation for surprising content (e.g., \textit{Mary is a surprising/boring person}), and whether the content is actually surprising (e.g., \textit{digging a hole with a fork/shovel}). 

Human results exhibit an interaction effect between informativity cue and situation typicality in their reading times. We introduce a decomposition of this interaction into three effects: the \cgreen{\textbf{facilitation effect}} (e.g., \textit{fork} becomes easier to process in a surprising context than in a boring context), the \cred{\textbf{inhibition effect}} (e.g., \textit{shovel} becomes harder to process in a surprising context than in a boring context), and the \cblue{\textbf{reversal effect}} (e.g., \textit{fork} is easier to process than \textit{shovel} in a surprising context). These effects show processing differences between a pair of conditions (Definitions in \S\ref{informativity_human}.).  



 %
 %

In our work, we pose three questions to compare how LLMs and humans process unpredictable but informative content: (i) Do LLMs exhibit informativity effects through word suprisal in a given sentence? (\S\ref{surprisal}) (ii) Are these effects reflected in models' predictive distributions? (\S\ref{sentence_completion}) (iii) Do these effects require explicit reasoning? (\S\ref{experiment3})

To answer these questions, we measure how much LLMs expect the situation (a)typical information in a context, calculate and compare the direction and relative size of informativity effects. We also test whether the effects vary with how explicitly the context raises informativity expectations. These contexts contain semantic or syntactic cues or communicative environment (see examples of context types in \autoref{studies}).  

Across three experiments, we find that
informativity effects found in humans are observed in LLMs only when the licensing context is semantically explicit. Under implicit cues (e.g., cleft constructions and conversational context), the effects are inconsistent across experiments. We also find that (i) LLM surprisal captures the \cgreen{\textbf{facilitation}} of unpredictable targets but does not capture the \cblue{\textbf{reversal}} effect, in which unpredictable continuations become easier to process than predictable ones when the context raises informativity. (ii) In a sentence completion task, models produce more variable and less typical continuations, as expected when the context calls for informative content. (iii) When probed for explicit judgments, reasoning models exhibit all three effects, suggesting that informativity effects in LLMs arise through post-hoc inference rather than being internalized in the predictive distribution. We also replicate these experiments on a larger dataset we constructed (\Cref{dataset_construction,expanded_results}).

In sum, our results suggest that under a surprisal account of processing effort, models’ next-word distributions cannot explain informativity effects found in human reading time. Post-hoc inference with discourse integration, however, can help them resolve informative content that conflicts with their predictive prior --- provided the context explicitly raises informativity expectations.

\section{Background}
We introduce how predictability and informativity interact in different stages of processing in humans (\S\ref{stages}) and the linguistic environments that give rise to such informativity expectations (\S\ref{informativity_cue}). 



\subsection{Predictability and Informativity in Sentence Processing}
\label{stages}
When comprehenders process sentences, they may construct a situation model of the events being described (e.g., \textit{digging a hole}) and pre-activate related concepts (e.g., \textit{dirt}, \textit{shovel}, \textit{gardening}) to facilitate processing \cite{federmeier1999rose,brothers2020going}. 
Reading times have shown that such predicted words are read more quickly, whereas unexpected ones (e.g., \textit{fork}) typically lead to a slowdown. This effect is measurable immediately on the (un)expected word (i.e., target). 
Both lexical retrieval of the target and its instantiating in the existing situation model may contribute to processing difficulty. \textbf{Informativity effects}, by contrast, have been measured at a later stage of processing, usually at the sentence-final position and not at the target word (we present this human finding in \S\ref{informativity_human}). This may be because a comprehender who encounters an unsurprising target word cannot know whether more surprising content will follow and satisfy the informativity expectation. We address this uncertainty by including wh-cleft constructions \cite{rohde2021s} in our experiments, which isolate the informativity effect at the target position (because the cleft marks where the informative content will appear; \S\ref{informativity_cue}). In that case, the informativity effect appears along with the predictability effect at the target word, and comprehenders may suppress the activation of typical content (e.g., \textit{digging a hole with a shovel}) at that position. 


\subsection{Informativity Cue}
\label{informativity_cue}
\begin{table}[t]
\small
\centering
\tabcolsep=4pt
\resizebox{\linewidth}{!}{%
\begin{tabular}{lll}
\toprule
 & \textbf{Context Cue} & \textbf{Example}               \\ \midrule
\makecell[l]{\citeauthor{rohde2021s} \\ (\citeyear{rohde2021s})}    & \makecell[l]{Unusual \\ protagonist} & Mary is a \textit{boring/ surprising} person. 
 \\ \cmidrule{2-3}
      & Wh-cleft & \makecell[l]{\textit{She was digging/ What she was digging} \\ the hole with...} 
\\ \cmidrule{2-3}
        & \makecell[l]{Intentionality\\ in conversation} & \makecell[l]{ 
        [In the format of a text-message conversation \\ where two speakers exchange intentionally.]
        \\
        \textbf{Amanda}: Buying a present for me?! \\ 
        \textbf{Joe}: ...price for the socks  was \textit{actually}...\\
       }
\\ \midrule 
\makecell[l]{\citeauthor{jiang2025expecting} \\ (\citeyear{jiang2025expecting})}  & \makecell[l]{Unrealistic \\ scenario}  & \makecell[l]{Mary is telling her friend what she \\\textit{did/ dreamt}.} \\ 
\bottomrule
\end{tabular}
}
\caption{A summary of informativity psycholinguistic studies we used. See full stimuli in \autoref{data_full}.}
\label{studies}
\end{table}
Speakers use contextual cues to signal that upcoming information may be newsworthy. We summarize the types of cues used in our experimental conditions in order of how \textit{implicit} they are in signaling informativity. See examples in \autoref{studies}.

\textbf{Lexical cues} can signal upcoming unexpected information  \cite{xiang2015reversing}. 
An example is the word \textit{surprising} as opposed to \textit{boring} \cite{rohde2021s}, or \textit{dream} as opposed to \textit{did} \cite{jiang2025expecting}. 

Speakers often use \textbf{syntactic constructions} to foreshadow discourse-new information, such as \textit{there}-construction \cite{ward2006information} and wh-cleft \cite{birner2004discourse}. Such constructions also signal the location of focused, often new, information denoted by the square in the examples (e.g., \textit{There is $\blacksquare$}; \textit{What ... is $\blacksquare$}). Thus, in the wh-cleft condition, informativity effects in humans emerge immediately at the target, unlike in the other cues.

Humans \textbf{communicate} in real life \textbf{intentionally}. Producing an utterance is costly for the speaker, and this cost is weighed against the utterance's utility—the informational value it adds to what the listener already believes \cite{frank2012predicting}. Therefore, one can expect newsworthy, informative messages in a conversation even without any explicit cues \cite{rohde2021s}.

\section{Human Findings and Experimental Design}
\label{task_datasets}

We first summarize informativity effects observed in humans from psycholinguistic studies (\S\ref{datasets}, \S\ref{informativity_human}), which serve as a point of comparison for our subsequent LLM experiments. We then describe the experimental design (\S\ref{task_design}). 
\begin{figure*}[htbp]
    \centering
    \includegraphics[width=0.76\linewidth]{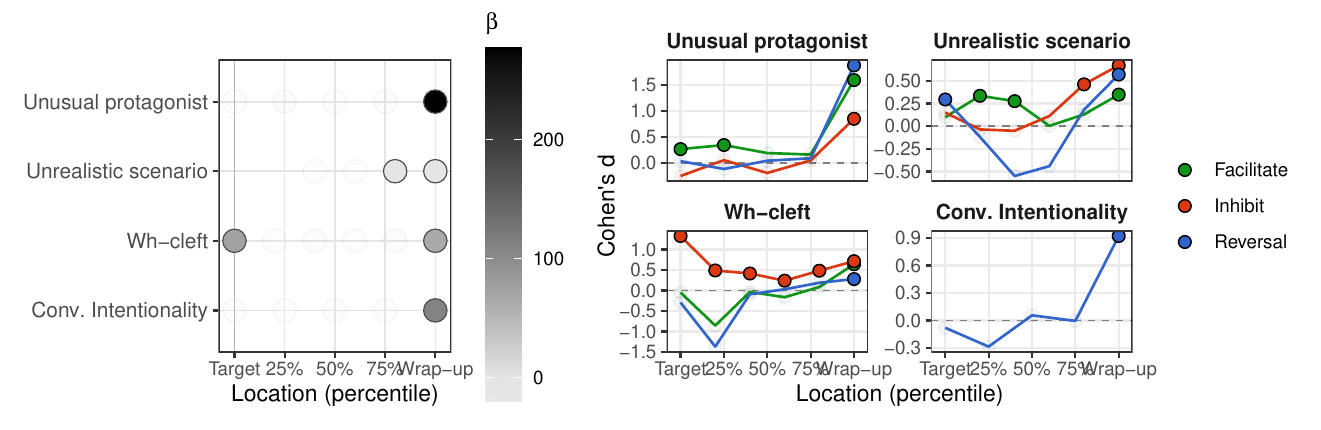}
    \caption{Findings from psycholinguistic studies of informativity  \cite{rohde2021s,jiang2025expecting}. \textbf{Left}: interaction effects coefficient $\beta$ is between cue and processing ease from linear mixed-effects models. \textbf{Right}: conditional informativity effects using standardized mean reaction time (ms) difference (transparent where $d<0.2$). Effect size is calculated according to definitions in \S \ref{informativity_human}, where a positive effect size is expected. Conv. Intentionality didn't have a $c_0$ condition in the human study.
    }
    \label{findings}
\end{figure*}
\subsection{Datasets} 
\label{datasets}
We adapt four human reading time experiments on informativity effects that vary in how explicitly the context raises informativity expectations (\autoref{studies}). Each experiment dataset is a $2 \times 2$ factorial study: the first variable $c = \{c_0, c_1\}$ is the \textbf{context}, whose cue respectively fails to raise or raises the expectation of an informative continuation (e.g., Mary is a \textit{boring/surprising} person), and the second variable $m = \{${$m_0$}, {$m_1$}$\}$ is the \textbf{situation typicality} (or predictability) \textbf{of the continuation}, whose values are respectively predictable and unpredictable (e.g., digging a hole with a \textit{shovel/ fork}). The location where the predictability of the continuation diverges is the \textbf{target word}. The setup constructs minimal pairs to factor out the effect of length. An unpredictable target word can be an atypical instrument for an action (e.g., chopping the carrots with a \textit{shovel}; \citet{brown1987adapting}), an uncommon location for an action (e.g., putting meat in the \textit{closet}), or an atypical value (e.g., \textit{\$150} socks).  See \autoref{data_full} for complete experimental conditions and stimuli. The conversation intentionality condition \cite{rohde2021s} is an exception to the dataset setup above, where $c=\{c_1\}$ and the original study doesn't include a control where the context fails to raise the informativity expectations. We here include the control $c_0$ that paraphrases dialogue into plain statements. 

\subsection{Informativity Effect in Humans} \label{informativity_human}
Previous psycholinguistic studies have measured the expectedness $E$ in humans using self-paced reading time, recorded at each word from the target to sentence-final. Shorter reading times indicate easier sentence processing, indicating that it is more expected. 
Human studies found a $2\times2$ interaction between the presence of an informativity cue and the presence of an informative target word \cite{rohde2021s,jiang2025expecting}. The different components of this interaction can be tested in terms of three informativity effects:
\begin{enumerate}[label=(\arabic*)]
    \item \cgreen{\textbf{Facilitation effect}}: Informativity expectations raised by context make the unpredictable target easier to process.
    \setlength{\abovedisplayskip}{4pt}
    \setlength{\belowdisplayskip}{4pt}
    \[E(c_1, m_1) - E(c_0, m_1) > 0\]
    \item \cred{\textbf{Inhibition effect}}: Informativity expectations raised by context make the predictable target harder to process, as it violates informativity expectations.
    \setlength{\abovedisplayskip}{4pt}
    \setlength{\belowdisplayskip}{4pt}
    \[E(c_0, m_0) - E(c_1, m_0) > 0\]
    \item \cblue{\textbf{Reversal effect}}: An unpredictable target is easier to process than a predictable one when the context raises informativity expectations, which reverses the effects where situation-predictable content is easier to process. 
    \[E(c_1, m_1) - E(c_1, m_0) > 0\]

\end{enumerate}


We define these effects to examine the context-driven change in the probability \textit{ranking} of situation a(typical) continuations. We leave quantitative relationships between these effects to future work.

\autoref{findings} summarizes the findings in these studies: the left figure displays the interaction effects between context and situation typicality, showing significant effects at the sentence wrap-up in all context types. The wh-cleft condition additionally shows a significant effect immediately at the target due to the cleft structure signaling where the informative content will appear, as predicted in \S\ref{informativity_cue}. 

While humans show informativity effects regardless of how explicit the context is, there are marked differences in effect size, with the unrealistic scenario condition exhibiting the weakest effect.
The \cblue{\textbf{reversal effect}} provides particularly strong evidence for informativity: experimental conditions with stronger interaction effects also show stronger \cblue{\textbf{reversal effects}}. 
Prior human studies do not all consistently find statistical significance for all three effects, but they do all show the expected effect directions at sentence wrap-up (\autoref{findings} right). 

As we will examine in \S\ref{surprisal}, near the target position, humans show a strong \cred{\textbf{inhibition}} effect in the wh-cleft context and a weak positive \cgreen{\textbf{facilitation}} effect in the unusual protagonist and unrealistic scenario context.


\subsection{Model Experiments} 
\label{task_design}
To examine whether LLMs exhibit informativity effects, our experiments measure how much LLMs expect the target in a context that raises informativity expectations.\footnote{We cannot directly correlate our measures with human reading times because that data is unavailable.} 
We evaluate each context-typicality combination separately to test whether previous human findings hold. LLMs receive only one item at inference without accumulating context to avoid learning effects where repeated exposure to informative utterances reduces models' sensitivity to what is situation-atypical. 




\begin{table*}[htbp]
\centering
\resizebox{0.78\textwidth}{!}{%
\begin{tabular}{llccc|ccc|ccc|ccc}
&& \multicolumn{3}{l}{$\leftarrow$ \textit{More explicit}} & & & \multicolumn{1}{c}{} & & & \multicolumn{1}{c}{} & \multicolumn{3}{r}{\textit{More implicit} $\rightarrow$ }  \\
\toprule
& & \multicolumn{3}{c}{\textbf{Unusual Protagonist}} & \multicolumn{3}{c}{\textbf{Unrealistic Scenario}} & \multicolumn{3}{c}{\textbf{Wh-cleft}} & \multicolumn{3}{c}{\textbf{Conv.\ Intentionality}} \\
\cmidrule(lr){3-5} \cmidrule(lr){6-8} \cmidrule(lr){9-11} \cmidrule(lr){12-14}
\textbf{Model} & \textbf{Type} & Fac. & Inh. & \multicolumn{1}{c}{Rev.} & Fac. & Inh. & \multicolumn{1}{c}{Rev.} & Fac. & Inh. & \multicolumn{1}{c}{Rev.} & Fac. & Inh. & Rev. \\
\midrule
Llama-3-8B & Base
  & \cellcolor{green!19}1.594 & \cellcolor{green!22}1.933 & \cellcolor{red!44}-4.353
  & \cellcolor{green!25}0.814 & \cellcolor{green!10}0.328 & \cellcolor{red!33}-8.872
  & \cellcolor{green!45}\textbf{0.880} & \cellcolor{red!7}-0.131 & \cellcolor{red!32}\textbf{-6.576}
  & \cellcolor{red!17}-1.234 & \cellcolor{green!19}2.047 & \cellcolor{red!10}-0.668 \\
Llama-3-8B & Instruct
  & \cellcolor{green!45}\textbf{3.786} & \cellcolor{green!31}2.725 & \cellcolor{red!30}-2.990
  & \cellcolor{green!45}\textbf{1.471} & \cellcolor{green!11}0.360 & \cellcolor{red!34}-9.118
  & \cellcolor{green!37}0.714 & \cellcolor{red!24}-0.471 & \cellcolor{red!42}-8.558
  & \cellcolor{red!15}-1.076 & \cellcolor{green!22}2.453 & \cellcolor{red!15}-1.014 \\
Llama-3.1-8B & Base
  & \cellcolor{green!19}1.623 & \cellcolor{green!20}1.784 & \cellcolor{red!45}-4.441
  & \cellcolor{green!26}0.841 & \cellcolor{green!4}0.139 & \cellcolor{red!34}-9.148
  & \cellcolor{green!21}0.418 & \cellcolor{red!17}-0.320 & \cellcolor{red!36}-7.467
  & \cellcolor{red!10}-0.744 & \cellcolor{green!15}1.598 & \cellcolor{red!13}-0.898 \\
Llama-3.1-8B & Instruct
  & \cellcolor{green!35}2.904 & \cellcolor{green!25}2.225 & \cellcolor{red!32}-3.148
  & \cellcolor{green!33}1.086 & \cellcolor{green!5}0.158 & \cellcolor{red!32}\textbf{-8.626}
  & \cellcolor{green!41}0.808 & \cellcolor{red!16}-0.305 & \cellcolor{red!35}-7.171
  & \cellcolor{red!8}-0.566 & \cellcolor{green!15}1.637 & \cellcolor{red!17}-1.207 \\
Llama-3.1-70B & Base
  & \cellcolor{green!24}2.019 & \cellcolor{green!30}2.636 & \cellcolor{red!35}-3.442
  & \cellcolor{green!25}0.832 & \cellcolor{green!6}0.190 & \cellcolor{red!35}-9.309
  & \cellcolor{red!45}-0.079 & \cellcolor{red!25}-0.490 & \cellcolor{red!43}-8.856
  & \cellcolor{red!45}-3.230 & \cellcolor{green!23}2.527 & \cellcolor{red!30}-2.094 \\
Llama-3.3-70B & Instruct
  & \cellcolor{green!43}3.647 & \cellcolor{green!45}\textbf{4.005} & \cellcolor{red!13}\textbf{-1.276}
  & \cellcolor{green!39}1.281 & \cellcolor{green!4}0.121 & \cellcolor{red!35}-9.413
  & \cellcolor{green!33}0.647 & \cellcolor{red!45}-0.867 & \cellcolor{red!45}-9.217
  & \cellcolor{red!26}-1.871 & \cellcolor{green!12}1.301 & \cellcolor{red!45}-3.113 \\ \midrule
Qwen2.5-7B & Base
  & \cellcolor{green!15}1.236 & \cellcolor{green!31}2.774 & \cellcolor{red!32}-3.137
  & \cellcolor{green!18}0.604 & \cellcolor{green!13}0.425 & \cellcolor{red!34}-9.011
  & \cellcolor{green!43}0.849 & \cellcolor{red!16}-0.311 & \cellcolor{red!34}-7.058
  & \cellcolor{red!29}-2.115 & \cellcolor{green!27}3.000 & \cellcolor{red!15}-1.047 \\
Qwen2.5-7B & Instruct
  & \cellcolor{green!27}2.308 & \cellcolor{green!36}3.245 & \cellcolor{red!31}-3.083
  & \cellcolor{green!39}1.288 & \cellcolor{green!3}0.092 & \cellcolor{red!45}-12.054
  & \cellcolor{green!28}0.546 & \cellcolor{red!43}-0.829 & \cellcolor{red!43}-8.873
  & \cellcolor{red!30}-2.146 & \cellcolor{green!45}\textbf{4.932} & \cellcolor{red!17}-1.207 \\
Qwen3-32B & Reason
  & \cellcolor{green!32}2.724 & \cellcolor{green!26}2.332 & \cellcolor{red!36}-3.596
  & \cellcolor{green!0}0.005 & \cellcolor{green!45}\textbf{1.497} & \cellcolor{red!37}-9.782
  & \cellcolor{green!40}0.779 & \cellcolor{green!45}\textbf{0.275} & \cellcolor{red!35}-7.108
  & \cellcolor{red!5}\textbf{-0.336} & \cellcolor{green!24}2.590 & \cellcolor{red!5}\textbf{-0.320} \\
\bottomrule
\end{tabular}
}
\caption{Informativity effects at target word as standardized surprisal differences across items. Positive values (green) indicate an expected effect found in the model, while negative values (red) indicate an unexpected effect. For each context and effect, boldface indicates the model with the largest effect size.}
\label{experiment1_target}
\end{table*}

\section{Experiment 1: Surprisal fails to align with informativity}
\label{surprisal}

To answer \textit{whether LLMs show informativity effects when processing a given sentence}, we first investigate model expectedness through surprisal. Surprisal is strongly supported by empirical studies of language processing, showing that words with higher surprisal often lead to longer reading times. 

Some informativity effects can be explained through word predictability \cite{futrell2012processing}: a context that raises informativity expectations will flatten the probability distribution over possible targets. As a result, a predictable target becomes less likely (\cred{\textbf{inhibition effect}}) while an unpredictable target becomes more likely (\cgreen{\textbf{facilitation effect}}). However, it is unclear whether such a context can reverse the predictability effect, so that an unpredictable target is more likely than a predictable one (\cblue{\textbf{reversal effect}}).    
To investigate to what extent informativity effects arise from next word prediction, we calculate the surprisal of the target and words after the target conditioned on its context, or $\text{surprisal} =-\log P(m | \text{context})$.    



\subsection{Experimental setup} We calculate the final layer surprisal of the target and words after the target up to sentence wrap-up to estimate informativity effect. As model surprisal measures word tokens directly, no spillover assumptions is needed. We test on base and instruct models of Llama3-8B, Llama3.1-8B, Llama3.1-70B, Qwen2.5-7B and reasoning model Qwen3-32B across context types. We compute the surprisal difference between each pair of conditions for three informativity effects defined in \S\ref{task_datasets}. We flip signs on surprisal differences (since surprisal is inversely related to expectedness $E$) so a positive difference indicates the expected direction for that effect. We report their standardized effect sizes. 

Since larger models' surprisal estimates are known to deviate from human reading behavior \cite{oh2023transformer, kuribayashi-etal-2024-psychometric}, we follow \citet{kuribayashi2025large} to compute next-word probabilities from internal layers (see \autoref{experiment1_layer}). This setting does not change our findings below.


\subsection{Results}
\label{exp1_results}

We first show that \textbf{models show general and context-dependent plausibility} following \citet{kauf2024log}. Following \autoref{informativity_example}, they are denoted
\[
\resizebox{0.86\linewidth}{!}{%
  $\begin{alignedat}{2}
    \text{\textbf{General}}: P(\text{shovel} \mid \text{Mary digs \ldots{} with})
      & \\ 
      > P(\text{fork} \mid \text{Mary digs \ldots{} with})
       \\
    \text{\textbf{Context}}: P(\text{fork} \mid \text{Mary is a \textit{surprising} \ldots{} with})
      & \\
      > P(\text{fork} \mid \text{Mary digs \ldots{} with})
  \end{alignedat}$%
}
\]
We verify that the models we selected are actually able to distinguish plausible from implausible targets when no influencing context is present (or \textbf{general plausibility}). 
We then show that an \textit{a priori} implausible target becomes more expected in the presence of a context that raises informativity expectations (or \textbf{context-dependent plausibility}). See \autoref{experiment1_supplementary} for full results.

We compare the informativity effects found in previous human studies (\autoref{findings}) with the effects in model surprisal at the target (\autoref{experiment1_target}). LLMs show a positive \cgreen{\textbf{facilitation effect}} --- matching that in humans --- in contexts that explicitly raise informativity expectations: unusual protagonist (1.983, [1.544, 2.417]) and unrealistic scenario (1.059, [.739, 1.379]). One-sided Satterthwaite tests show positive effects in both contexts (adjusted $p<$.001). However, models show mostly negative \cred{\textbf{inhibition effects}} in the wh-cleft context, while these effects are strongly positive in humans. 
The result suggests that \textbf{the effects found in LLMs do not always align with those in humans.}
\begin{figure}[t]
    \centering
    \includegraphics[width=\linewidth]{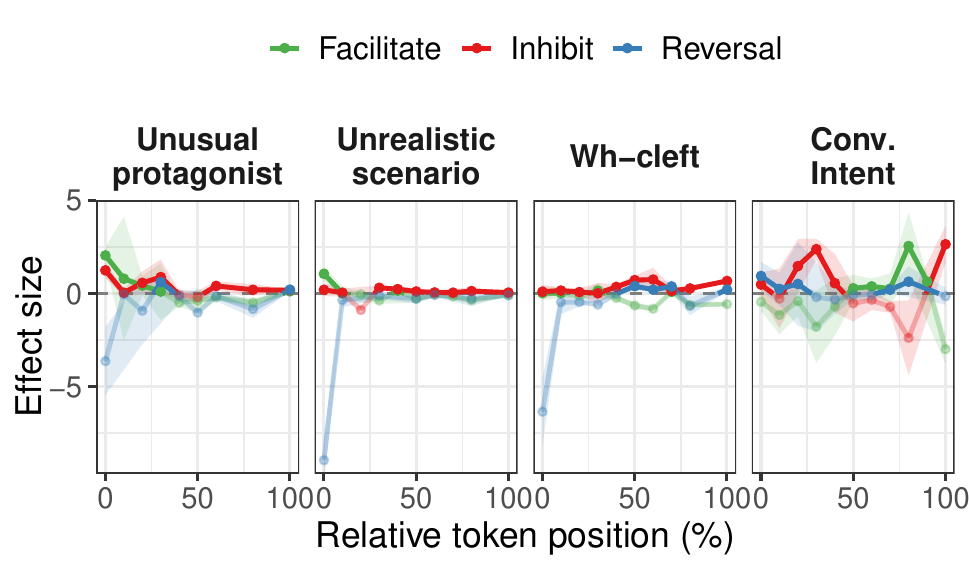}
    \caption{Informativity effect at words from the target to sentence wrapup using standardized surprisal differences. Positive values indicate an expected effect found in models, while negative values indicate an unexpected effect.}
    \label{wrapup}
\end{figure}

The most noticeable result is that \textbf{model surprisal fails to capture the} \cblue{\textbf{reversal effect}} \textbf{consistently} at the target (-4.593, [-5.412, -3.799]), or even at words after the target (\autoref{wrapup}), as evident in negative and near-zero effect sizes. One-sided Satterthwaite tests show negative reversal effects at in all contexts (adjusted $p<$.001). This finding is consistent with the constraint of expectation-based comprehension: comprehenders cannot predict what specific atypical instrument Mary would use to dig a hole, as there is an infinite set of atypical instruments besides \textit{fork} that would be appropriate to the discourse, all of which should be more expected than a typical instrument like \textit{shovel} for human comprehenders. But for a next word predictor no single atypical word become more probable than \textit{shovel}. The results thus suggest that \textbf{surprisal does not align with informativity effects} and challenge the hypothesis that predictability is linked to human cognitive processing effort.

We also find that, at the target, both facilitation and inhibition effects emerge in model surprisal only for the more explicit contexts (unusual protagonist and unrealistic scenario), as evident in positive effect sizes in \autoref{experiment1_target} and most strongly for the unusual protagonist context.

\section{Experiment 2: Model distributions expect informative continuations}
\label{sentence_completion}

In this experiment, we examine whether model expectations are modulated by licensing context using a sentence-completion task, which allows us to examine the full predictive distribution of a model. We confirm our account in \S \ref{exp1_results} that under informativity expectations, the  predictive distribution becomes flatter and less typical.


\subsection{Experimental setup}

We follow the procedures in \citet{reksnes2024tell} and \citet{jiang2025expecting} for our sentence completion task: we use the same experimental items as in the previous experiment and include all contexts except Conversation Intentionality. Items in these contexts end with a noun phrase (e.g., \textit{using a fork}), which simplifies our evaluation. We ask the models to predict the target word. 

We run the sentence completion task on LLMs using (1) \textbf{direct} measurement of word probabilities from the models’ internal logits over the vocabulary and (2) word probabilities after \textbf{meta}linguistic prompting using the same instruction given to humans in prior work (see prompts in \autoref{verbalized_expectedness_prompts}).\footnote{See \citet{hu2023prompting} for a discussion of both methods.} We use base and instruct models of Llama3.1-8B and Qwen2.5-7B as well as reasoning model Qwen3-32B to examine whether post-training shifts target distribution. As we are interested in testing informativity effects in their predictive distribution, we use multinomial sampling based on internal logits over vocabulary for up to two words ($N = 100$) to truncate the completion to a single noun phrase. 
We exclude completions that do not contain a complete noun phrase. We standardize sentence completions by collapsing noun phrases that refer to the same object but differ in case, punctuation, spelling, and initial article. 




\subsection{Evaluation}
We expect that sentence completions are less typical and more variable in contexts that raise informativity expectations than those that don't. We obtain typicality and variability measures of completions as follows: we judge the real-world \textbf{typicality} of objects mentioned in completion when no influencing context is present. Following \citet{reksnes2024tell}, we conduct a pre-test using GPT 5.4 and Claude Sonnet 4.6 to identify a set of completions ($N=50$ respectively) one would be likely to find in the scenario. We evaluate each completion using a typicality score, which is the proportion of pre-test completions that mention that noun with add-one smoothing.  

We obtain the following metrics for response \textbf{variability}: (1) \textbf{Tok}: the entropy of next token distribution; (2) \textbf{Comp}: the entropy of top $k$ completions standardized\footnote{An equal sample size $k$ ensures that entropy scores are not distorted. }; (3) \textbf{Rel}: the relative entropy, or Kullback-Leibler (KL) divergence, of completions between $c_1$, the condition that raises informativity expectations, and $c_0$, the condition that fails to raise such expectation. We also compare completion entropy and typicality between human and LLM cloze in surprising protagonist and unrealistic scenario conditions.


\begin{table*}[ht]
\centering
\resizebox{0.78\textwidth}{!}{%
\begin{tabular}{llcccc|cccc|cccc}
& & \multicolumn{3}{l}{$\leftarrow$ \textit{More explicit}} & \multicolumn{1}{c}{} & & & & \multicolumn{1}{c}{} & & \multicolumn{3}{r}{\textit{More implicit} $\rightarrow$ }  \\
\toprule
& & \multicolumn{4}{c}{\textbf{Unusual Protagonist}} & \multicolumn{4}{c}{\textbf{Unrealistic Scenario}} & \multicolumn{4}{c}{\textbf{Wh-cleft}} \\
\cmidrule(lr){3-6} \cmidrule(lr){7-10} \cmidrule(lr){11-14}
& & \multicolumn{3}{c}{\textbf{Var} ($\uparrow$)} & \multicolumn{1}{c}{} & \multicolumn{3}{c}{\textbf{Var} ($\uparrow$)} & \multicolumn{1}{c}{} & \multicolumn{3}{c}{\textbf{Var} ($\uparrow$)} & \\
\cmidrule(lr){3-5} \cmidrule(lr){7-9} \cmidrule(lr){11-13}
\textbf{Model} & \textbf{Method} & Tok & Comp & Rel & \multicolumn{1}{c}{\textbf{Typ} ($\downarrow$)} & Tok & Comp & Rel & \multicolumn{1}{c}{\textbf{Typ} ($\downarrow$)} & Tok & Comp & Rel & \textbf{Typ} ($\downarrow$) \\
\midrule
\multirow{2}{*}{Llama-3.1-8B}          & Direct & \cellcolor{green!27}1.67 & \cellcolor{green!37}1.16 & \cellcolor{green!46}11.81 & \cellcolor{green!13}-0.86 & \cellcolor{green!87}0.75 & \cellcolor{green!84}0.68 & \cellcolor{green!12}5.73  & \cellcolor{green!58}-0.12 & \cellcolor{green!34}0.67  & \cellcolor{green!84}0.54  & \cellcolor{green!71}13.09 & \cellcolor{green!45}-0.43 \\
                                       & Meta   & \cellcolor{green!26}1.62 & \cellcolor{green!35}1.09 & \cellcolor{green!42}11.65 & \cellcolor{green!25}-1.03 & \cellcolor{green!68}0.61 & \cellcolor{green!60}0.47 & \cellcolor{green!46}6.46  & \cellcolor{green!71}-0.15 & \cellcolor{red!24}-0.46   & \cellcolor{green!40}0.21  & \cellcolor{green!42}11.23 & \cellcolor{red!32}0.26   \\
\multirow{2}{*}{Llama-3.1-8B-Instruct} & Direct & \cellcolor{green!33}1.85 & \cellcolor{green!42}1.31 & \cellcolor{green!69}12.84 & \cellcolor{green!38}-1.23 & \cellcolor{green!83}0.72 & \cellcolor{green!88}0.72 & \cellcolor{green!72}7.03  & \cellcolor{green!88}-0.19 & \cellcolor{green!40}0.77  & \cellcolor{green!51}0.29  & \cellcolor{green!69}12.97 & \cellcolor{green!31}-0.25 \\
                                       & Meta   & \cellcolor{green!33}1.85 & \cellcolor{green!54}1.67 & \cellcolor{green!88}13.70 & \cellcolor{green!12}-0.84 & \cellcolor{green!88}0.76 & \cellcolor{green!61}0.48 & \cellcolor{green!88}7.37  & \cellcolor{green!54}-0.11 & \cellcolor{red!31}-0.60   & \cellcolor{red!88}-0.57   & \cellcolor{green!78}13.52 & \cellcolor{green!33}-0.28 \\ \midrule
\multirow{2}{*}{Qwen2.5-7B}            & Direct & \cellcolor{green!56}2.59 & \cellcolor{green!43}1.34 & \cellcolor{green!12}10.32 & \cellcolor{green!42}-1.28 & \cellcolor{green!80}0.70 & \cellcolor{green!67}0.53 & \cellcolor{green!29}6.10  & \cellcolor{green!50}-0.10 & \cellcolor{red!12}-0.23   & \cellcolor{green!61}0.37  & \cellcolor{green!19}9.83  & \cellcolor{red!14}0.03   \\
                                       & Meta   & \cellcolor{green!72}3.11 & \cellcolor{green!47}1.46 & \cellcolor{green!25}10.88 & \cellcolor{green!57}-1.51 & \cellcolor{green!63}0.57 & \cellcolor{green!56}0.43 & \cellcolor{green!35}6.23  & \cellcolor{red!12}0.01    & \cellcolor{red!27}-0.52   & \cellcolor{red!25}-0.10   & \cellcolor{green!67}12.84 & \cellcolor{red!22}0.14   \\
\multirow{2}{*}{Qwen2.5-7B-Instruct}   & Direct & \cellcolor{green!88}3.63 & \cellcolor{green!88}2.72 & \cellcolor{green!30}11.14 & \cellcolor{green!88}-1.96 & \cellcolor{green!16}0.22 & \cellcolor{green!34}0.24 & \cellcolor{green!64}6.85  & \cellcolor{green!71}-0.15 & \cellcolor{red!29}-0.56   & \cellcolor{red!28}-0.12   & \cellcolor{green!46}11.49 & \cellcolor{red!21}0.12   \\
                                       & Meta   & \cellcolor{green!35}1.91 & \cellcolor{green!49}1.51 & \cellcolor{green!76}13.17 & \cellcolor{green!24}-1.01 & \cellcolor{green!49}0.47 & \cellcolor{green!75}0.60 & \cellcolor{green!50}6.56  & \cellcolor{green!46}-0.09 & \cellcolor{red!88}-1.72   & \cellcolor{red!68}-0.42   & \cellcolor{green!88}14.15 & \cellcolor{green!88}-0.97 \\
\multirow{2}{*}{Qwen3-32B}             & Direct & \cellcolor{green!18}1.38 & \cellcolor{green!25}0.80 & \cellcolor{green!55}12.25 & \cellcolor{green!34}-1.16 & \cellcolor{green!12}0.19 & \cellcolor{red!12}-0.04  & \cellcolor{green!64}6.85  & \cellcolor{green!67}-0.14 & \cellcolor{green!18}0.35  & \cellcolor{green!41}0.22  & \cellcolor{green!12}9.37  & \cellcolor{green!64}-0.67 \\
                                       & Meta   & \cellcolor{green!12}1.17 & \cellcolor{green!12}0.39 & \cellcolor{green!80}13.33 & \cellcolor{green!41}-1.27 & \cellcolor{green!17}0.23 & \cellcolor{green!34}0.24 & \cellcolor{green!61}6.78  & \cellcolor{red!46}0.09    & \cellcolor{green!15}0.29  & \cellcolor{red!12}0.00    & \cellcolor{green!46}11.48 & \cellcolor{red!12}-0.01   \\
\bottomrule
\end{tabular}%
}
\caption{Standardized mean difference for the variability (\textbf{Var}) and typicality (\textbf{Typ}) of completions across items between informative and uninformative contexts ($c_1 - c_0$). Green cells indicate expected effect sizes, and red cells unexpected.}
\label{experiment2}
\end{table*}

\begin{table}[t]
\centering
\resizebox{\linewidth}{!}{%
\small
\begin{tabular}{@{}lrrrr|rr@{}}
\toprule
 &  \multicolumn{4}{c|}{\textbf{Comp}} & \multicolumn{2}{c}{\textbf{Typ}} \\
\addlinespace
 & \multicolumn{2}{c}{\textbf{Unusual Protagonist}} & \multicolumn{2}{c|}{\textbf{Unrealistic Scenario}} & \multicolumn{2}{r}{\textbf{Unrealistic Scenario}} \\
\cmidrule(lr){2-3} \cmidrule(lr){4-5} \cmidrule(lr){6-7}
 & Human & LLM & Human & LLM & Human & LLM \\
\midrule
$c_0$ & 1.62 & 1.77 & 1.38 (.59) & 1.46 (.38) & .565 (.155) & .524 (.175) \\
$c_1$ & 2.56 & 2.04 & 1.77 (.59) & 1.59 (.42) & .521 (.137) & .511 (.167) \\
$p$  & \textbf{< .001} & \textbf{< .001} & \textbf{< .001} & \textbf{< .01} & \textbf{< .05} & 	.534 \\
\bottomrule
\end{tabular}
}
\caption{Human and LLM cloze completion entropy and typicality in
uninformative ($c_0$) and informative ($c_1$) contexts. Standard deviations in parenthesis. $p$ indicates permutation test statistics for mean differences.}
\label{tab:human-llm-cloze}
\end{table}
\subsection{Results}

\autoref{experiment2} shows the effect size of variability and typicality metrics between the cue that raises informativity and that does not, or $c_1 -c_0$. See \autoref{completion_examples} for completion examples. As expected, models that expect informativity exhibit positive effect sizes in all variability metrics and negative effect sizes in typicality (expected effects marked as green cells)\footnote{Effects are negative for typicality (-.075 [-.090, -.060]) and positive for token (.546 [.478, .615]), completion (.312 [.260, .363]), and relative entropy (10.189 [9.820, 10.554]).}. Informative contexts increase completion entropy significantly in both humans and LLMs (\autoref{tab:human-llm-cloze}). However, typicality decreases significantly in informative contexts only in humans ($\beta = -.044, p<.05$), while LLMs exhibit the same trend ($\beta = -.013, p=.256$)\footnote{We use a mixed-effects model of typicality with context as a fixed effect and participant (or model) and item random effects.}.

In addition, \textbf{the informativity effect is more consistent with more explicit cues for informativity} (unrealistic scenario and unusual protagonist), as evident in less unexpected effect sizes (marked as red cells) across models. Noticeably, most models fail to expect informative continuations in the wh-cleft as there is no lexical cues inducing such expectation semantically (\S\ref{experiment3}), where the 95\% credible intervals for typicality, token entropy, and completion entropy all include zero.

In the unusual protagonist context, instruction-tuned models overall produce more variable and atypical completions (i.e., increase in \textbf{Comp} and decrease in \textbf{Typ}) under informativity expectations across models and methods. In summary, our results indicate that LLMs are sensitive to contexts that explicitly raise informativity expectations and expect more informative continuations in sentence completion. While Experiment 1 shows the typical target remains more probable in informative contexts, Experiment 2 shows model distributions spread across more licensed atypical continuations in informative contexts.



\section{Experiment 3: Post-hoc inferences align with
informativity}
\label{experiment3}

Prior work provides evidence that informativity effects in humans may not show up until later stages of processing \cite{rohde2021s, xu2025informativity}, which raises the question of whether there exists some mechanism that handles unpredictable but informative utterances. Reasoning in language models has been shown to improve performance on inference-intensive language tasks. Our task may involve inferences on (1) whether the context raises informativity expectations, (2) how typical are events mentioned in the provided context, and (3) how do given events compare with some alternative events in the context, which raises our last question:  \textit{Do LLMs require explicit reasoning to resolve informativity expectations?} To test this, we probe reasoning models for how well an unpredictable target fits the context and examine whether they support previous human findings.

\subsection{Method}
We measure how much an influencing context shifts the expectedness of the target. To this end, we take inspiration from \textit{Bayesian surprise} \cite{itti2009bayesian, agarwal2026autodiscovery}. While surprisal measures how unexpected an observation is given a \textit{current} model, Bayesian surprise measures the extent to which that observation \textit{updates} the model’s prior beliefs \cite{zarcone2016salience}. In our task, the observation is the context that raises informativity expectation. Consider a model with a prior distribution $P(m)$ for how much they expect a given target $m$ when no influencing context is present. When the model observes an influencing context $c$, it adjusts the belief to posterior distribution $P(m|c)$. We define expectedness score as the signed Wasserstein distance between the posterior and prior distributions:
\[
\begin{aligned}
E(c,m) &= \operatorname{sgn}\!\left(\mathbb{E}_{X \sim P(m \mid c)}[X] - \mathbb{E}_{X \sim P(m)}[X]\right) \\
&\quad \cdot W_1\!\left(P(m \mid c), P(m)\right)
\end{aligned}
\]

In order to empirically estimate the prior and posterior distributions of the target, we sample expectedness scores from reasoning language models using verbalized numerical estimates \cite{tian-etal-2023-just} with and without the informativity cue (the premise control condition in \autoref{data_full}), and compute the three informativity effects over $E(c,m)$. Specifically, we ask the model to reason about how much it expects the target given its context and return a score between 0 and 100 indicating the level of expectedness. 






\label{inference_results}
\subsection{Experimental setup}
We calculate the verbalized expectedness score to estimate informativity effect after model reasoning. We adapt prompts from \citet{kamath2026humans} for probabilistic inference to probe how well the target fits the context (see prompts in \autoref{verbalized_expectedness_prompts}). Considering that previous human studies reduce priming effects by showing participants only a few stimuli \cite{rohde2021s}, we evaluate the model zero-shot. We test GPT-5, Qwen 32B and Qwen 235B with varying reasoning effort. For details on model and how we run inference, see \autoref{exp3_inference_detail}. 

\begin{figure}[t]
    \centering
    \includegraphics[width=\linewidth]{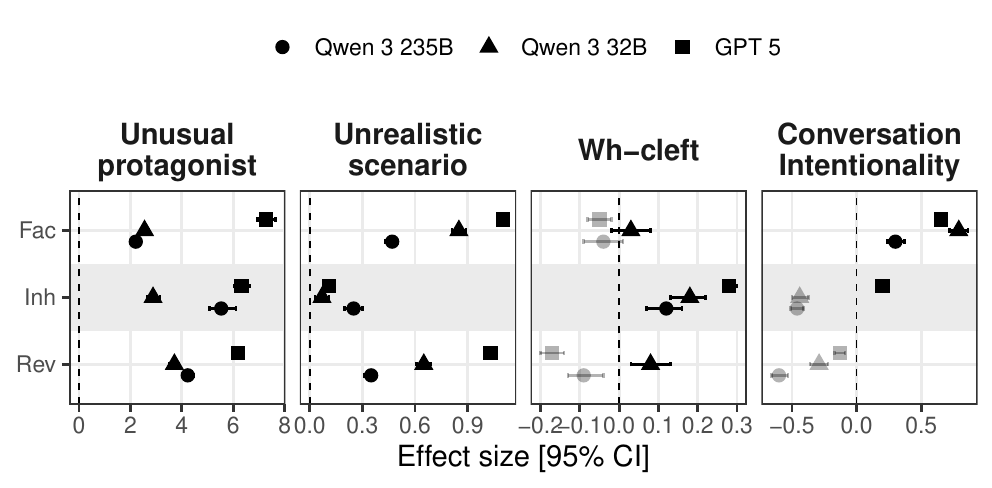}
    \caption{Standardized mean difference of verbalized expectedness for informativity effects with $95\%$ confidence interval. Positive effect sizes are expected.}
    \label{experiment3_effect_size}
\end{figure}

\begin{table}[t]
\centering
\resizebox{1\columnwidth}{!}{%
\begin{tabular}{lrrrrrr}
\toprule
& \multicolumn{2}{c}{\textbf{Facilitate}} & \multicolumn{2}{c}{\textbf{Inhibit}} & \multicolumn{2}{c}{\textbf{Reversal}} \\
\cmidrule(lr){2-3} \cmidrule(lr){4-5} \cmidrule(lr){6-7}
\textbf{Experiments} & Stat. & $p$ & Stat. & $p$ & Stat. & $p$ \\
\midrule
Unusual protagonist       & 25.34 & $\textbf{<0.001}$ & 26.45 & $\textbf{<0.001}$ & 23.77 & $\textbf{<0.001}$ \\
Unrealistic scenario      & 64.25 & $\textbf{<0.001}$ & 12.35 & 0.006           & 47.55 & $\textbf{<0.001}$ \\
Wh-cleft                  & 7.62  & 0.055           & 4.99  & 0.172           & 11.31 & 0.010 \\
Conv. Intentionality & 10.05 & 0.018           & 3.45  & 0.327           & 4.05  & 0.256 \\
\bottomrule
\end{tabular}
}
\caption{
Friedman Test results of informativity effect sizes across reasoning effort.}
\label{reasoning_effort_effect_size}
\end{table}
\subsection{Results}
While model surprisal fails to exhibit \cblue{\textbf{reversal effects}} observed in humans (\S\ref{surprisal}), these effects do emerge, and more strongly, through post-hoc inferences (\autoref{experiment3_effect_size}), indicated by positive effects found in all but the conversational intentionality context. Nevertheless, \textbf{informativity effects arise from more explicit cues}. None of the reasoning models show the \cblue{\textbf{reversal effects}} in the conversation intentionality context. The unusual protagonist context produces the strongest effects of all context types, likely because it raises informativity expectations more explicitly and induces a strong bias toward violations of real-world knowledge (e.g., a protagonist who \textit{never} does things the way you’d expect). Wh-cleft context exhibits the largest variance in effect size across items, which may be due to models failing to recognize cleft structure as a cue to raise informativity expectation (\S \ref{inferences_informativity}). We also find that reasoning effort doesn't significantly affect informativity effect size when a context raises informativity expectations more implicitly (\autoref{reasoning_effort_effect_size}).

\subsection{Informativity Inferences} 
\label{inferences_informativity}
To answer whether models expect informativity for the right reasons, we sample reasoning summaries in the critical condition ($c_1, m_1$) across items and contexts. We find shared inference patterns in these reasoning summaries and define the inference steps in \autoref{steps}. See additional analyses in \autoref{additional_inference_analyis}.

\begin{table}[h]
\centering
\small
\resizebox{\linewidth}{!}{%
\renewcommand{\arraystretch}{1.4}
\begin{tabular}{@{}p{2cm} p{3.5cm} p{3cm}@{}}
\toprule
\textbf{Step} & \textbf{Description} & \textbf{Example} \\
\midrule
\textbf{Cue Detection} &
The model picks up a cue in the context (\S\ref{informativity_cue}) that shapes how \textbf{expected} or \textbf{surprising} the target will be. &
The example describes my cousin Mary as \hl{surprising} and \hl{using unexpected methods}. \\
\addlinespace
\textbf{Target \mbox{Typicality}} &
The model judges how expected the target is within its sentence using world knowledge, before factoring in the cue from Step 1. &
An \hl{unusual} tool like a fork seems \hl{unexpected initially}. \\
\addlinespace
\textbf{Target \mbox{Alternatives}} &
The model names one or more alternatives for the target and compares them to the target. &
Normally, tools like \hl{a shovel} or \hl{spade} come to mind for that task. \\
\addlinespace
\textbf{Target Fit} &
The model brings evidence from the previous step and judges whether the target fits the overall context. &
With her surprising nature, (the fork) actually becomes a \hl{plausible} choice. \\
\addlinespace
\textbf{Accommodate} &
The model draws additional inferences that are not directly inferrable from the context to justify why a surprising target still fits. &
\hl{Maybe a child or someone limited in resources} might use (the fork). \\
\bottomrule
\end{tabular}
}
\caption{Informativity Inference Steps}
\label{steps}
\end{table}

Recent work has questioned whether reasoning traces faithfully reflect models’ internal computations \cite{kambhampati2025position}. We focus on a separate, underexplored question concerning the semantic validity of these traces in pragmatic tasks. Specifically, we examine whether the prominence of inferential steps in the traces signals when models fail to resolve informativity expectations. To do so, we annotate inference steps in 760 reasoning summaries with Claude Opus 4.8. We also subsample 30 summaries for human annotations by four participants. The mean inter-annotator agreement between human annotators and LLM is $\kappa = 0.79$. Annotation details are in \autoref{annotation}.

\begin{table}[t]
\centering\small
\setlength{\tabcolsep}{3.5pt}
\resizebox{0.97\columnwidth}{!}{%
\begin{tabular}{@{}lcccc@{}}
\toprule
                      & \textbf{Unus.}  & \textbf{Unreal.} & \textbf{Wh-}   & \textbf{Conv.}  \\
                      & \textbf{Prot.}  & \textbf{Scen.}   & \textbf{Cleft} & \textbf{Intent.} \\
\midrule
No cue / Cue points to \textbf{expected}   & \cellcolor{black!4}0.06 & \cellcolor{black!13}0.19 & \cellcolor{black!69}\textcolor{white}{0.99} & \cellcolor{black!39}\textcolor{white}{0.56} \\
Cue points to \textbf{surprising} $\checkmark$ & \cellcolor{black!66}\textcolor{white}{0.94} & \cellcolor{black!56}\textcolor{white}{0.80} & \cellcolor{black!1}0.01 & \cellcolor{black!31}\textcolor{white}{0.44} \\
\midrule
\addlinespace[2pt]
Typicality not judged / Target is \textbf{typical}   & \cellcolor{black!5}0.07 & \cellcolor{black!4}0.06 & \cellcolor{black!1}0.01 & \cellcolor{black!16}0.23 \\
Target is \textbf{atypical} $\checkmark$ & \cellcolor{black!65}\textcolor{white}{0.93} & \cellcolor{black!67}\textcolor{white}{0.95} & \cellcolor{black!69}\textcolor{white}{0.99} & \cellcolor{black!54}\textcolor{white}{0.77} \\
\midrule
\addlinespace[2pt]
No alterative              & \cellcolor{black!36}\textcolor{white}{0.52} & \cellcolor{black!25}0.36 & \cellcolor{black!11}0.15 & \cellcolor{black!50}\textcolor{white}{0.71} \\
Alterative more typical than target         & \cellcolor{black!23}0.33 & \cellcolor{black!45}\textcolor{white}{0.64} & \cellcolor{black!60}\textcolor{white}{0.85} & \cellcolor{black!19}0.27 \\
\midrule
\addlinespace[2pt]
Target doesn't fit the context   & \cellcolor{black!6}0.08 & \cellcolor{black!40}\textcolor{white}{0.57} & \cellcolor{black!62}\textcolor{white}{0.88} & \cellcolor{black!22}0.31 \\
Target fits the context $\checkmark$ & \cellcolor{black!64}\textcolor{white}{0.92} & \cellcolor{black!30}0.43 & \cellcolor{black!8}0.12 & \cellcolor{black!48}\textcolor{white}{0.69} \\
\midrule
\addlinespace[2pt]
No accommodation              & \cellcolor{black!38}\textcolor{white}{0.54} & \cellcolor{black!32}{0.45} & \cellcolor{black!28}0.40 & \cellcolor{black!16}0.23 \\
Attempted accommodation       & \cellcolor{black!32}{0.46} & \cellcolor{black!39}\textcolor{white}{0.55} & \cellcolor{black!42}\textcolor{white}{0.60} & \cellcolor{black!54}\textcolor{white}{0.77} \\
\bottomrule
\end{tabular}
}
\caption{Inference step distribution across models. $\checkmark$ indicates the correct inference for that step.}
\label{inference_distribution}
\end{table}
We show the distribution of inference steps for each context type in \autoref{inference_distribution}. A correct inference should find the cue $c_1$ that points to a surprising continuation, judge the target $m_1$ to be atypical, and conclude that the target overall fits the context (marked $\checkmark$). These reasoning chains help explain why \textit{informativity effects are affected by how explicitly the context licenses informativity}: LLMs fail to incorporate the cleft structure and conversational intentionality as contextual cues in their intermediate traces, or do not recognize them as cues that license a more surprising continuation. Although LLMs reliably judge the unpredictable target as atypical before considering the influencing context, in the wh-cleft condition they go on to verbalize that the target does not fit the overall context in 88\% of cases (\autoref{inference_distribution}). LLMs also find the unpredictable target unfit in the unrealistic scenario context the  majority of the time, which shows that reasoning models remain bounded by their world knowledge prior in the context of dreams.  

\section{Discussion}
Our experiments show that LLMs' predictive distributions cannot capture human informativity effects consistently across context types. Surprisal estimated by these models reflects error in lexical prediction. When the informativity expectations are raised, prediction error for the target word is likely high, making surprisal unfit to capture the processing ease of an unpredictable continuation. Reasoning LLMs can better resolve informative continuations that conflict with their predictive prior through post-hoc inferences. However, their ability depends on whether they can pick up cues that raise informativity expectations --- making implicit cues, such as wh-cleft and conversation intentionality, harder to comprehend. The finding reveals a pitfall in how LLMs may process information differently from humans: we routinely integrate highly unpredictable information (e.g., a rare name or a specific year) 
even in the absence of explicit semantic cues that facilitate comprehenders to expect newsworthy information.

We would also like to put our findings in the context of a second setting in which predictable linguistic units become difficult to process, as discussed in the introduction. LLMs also process highly predictable information (e.g., John \textit{paid} the cashier) differently than humans do. While humans adjust their situation prior in response to such underinformative utterances to repair the dip in information utility (e.g., perhaps John usually shoplifts, \citealp{kravtchenko2022informationally}), LLMs fail to react to the redundancy as an inappropriate contribution to the discourse \cite{kurch2024large}. Taken together, an informativity-driven model should therefore capture human processing patterns in settings where situation typicality may not directly translate to utterance felicity. To do so, it needs not only the probability of upcoming information but also the probability that a speaker would produce that information at all. It remains an open question, however, how to model the likelihood that human speakers produce utterances that adhere to communicative goals such as informativity.

\section*{Limitations}

We acknowledge several limitations that fall outside the scope of our findings and point toward future directions.

\paragraph{Comparison with Human Reading Time} Our experiments make direct comparison with informativity effects found in human reading time, which is an online measure of processing ease with low task demand \cite{hu2024auxiliary}. Our surprisal results at the target align with human findings at the target, especially for the facilitation effect. However, some expectedness measures in this work are offline metalinguistic judgments with additional task demand (i.e., interpreting the question and producing the answer), since an equivalent measure to reading time is unavailable. Future work can investigate how LLMs expect informativity internally, beyond surprisal, using other methods such as linear probing.  

\paragraph{Implication for Post-hoc Inferences} While Experiment 3 offers insight on what inference steps LLMs use to resolve informativity expectations, their implications on human comprehenders are less clear. LLMs may make overinferences to arrive at their conclusions compared to humans.   

\paragraph{Human Reference for Sentence Completion Task} In Experiment 2, we use a reference LLM different from the models we tested for the typicality pretest. While we believe the reference model can reliably approximate a set of typical completions without the influencing context, future studies can collect human completions to further examine how they differ from LLM completions.  
\paragraph{Minimal Pair Design} Although we reuse minimal pairs in previous psycholinguistic studies, future work could develop even better stimuli to yield results that are better controlled. For instance, uninformative targets in the unusual protagonist context do violate not only informativity expectation but also discourse coherence, which may conflate the \cblue{\textbf{reversal effect}} with incoherence. This may explain why LLMs exhibit informativity effects more strongly in the unusual protagonist context. In the unrealistic scenario context, the informativity expectations established early in the context may be attenuated by a series of typical, uninformative events that follow. Also, the manipulation could more carefully select plausible dream events as its target. 
While the $c_0$ condition in the unusual protagonist context induces comprehenders not to expect informative content (e.g., Mary is \textit{boring}), this constraint is less strong in the unrealistic scenario condition. 

\paragraph{Quantitative Relationships of Effects} While our experiment focuses on situations where their processing differences (without context) are large enough for informativity effects to be meaningful, we suspect that informativity effects can be influenced by the situation typicality of  experimental items. Future work may benefit from item-level analyses and examine the quantitative relationships of three effects.

\section*{Acknowledgments}

This work is funded by the Deutsche Forschungsgemeinschaft (DFG, German Research Foundation) -- GRK 2853/1 “Neuroexplicit Models of Language, Vision, and Action” - project number 471607914.



\bibliography{custom}

\newpage 

\appendix
\appendix
\crefalias{section}{appendix}
\section{Code and Materials}
Code and materials for experiments can be found at \url{https://anonymous.4open.science/r/llm_informativity}.
\section{Related Work}

\paragraph{Context Sensitivity} Our work on informativity effects is distinct from related work on LLM context sensitivity, where global context overrides comprehenders' default expectations by shifting to a different set of words \cite{kauf2024log}. In the example below, the context establishes that the girl is dressed as a bird, overriding comprehenders' prior expectations about how a girl is described. The context thereby makes the set of words describing a canary more predictable.  

\begin{quote}
    \small 
    The girl dressed up as a canary for Halloween. The girl had a little beak and a bright yellow tail.
\end{quote}

In our experiments, the global context suggests to comprehenders that a \textit{surprising} word will be used. However, it is difficult for comprehenders to predict which set of surprising words will be used. Therefore, comprehenders' expectations shift from a small set of predictable words to an unbounded set of unpredictable words \cite{futrell2012processing}.   

\paragraph{Informativity Effects in LLMs} While \citet{hao2023can} also investigated whether language model surprisal can explain the informativity bias in humans, they tested only unusual-protagonist contexts and smaller language models, whose size may affect their pragmatic capabilities and contextual sensitivity. In addition, they only report surprisal at the target. They find that surprisal using GPT-2, GPT-2-L, and GPT-3 fails to capture human-like patterns in informativity effects, which agrees with our findings.

\paragraph{Informativity Effects in Humans} This study focuses on informativity studies on \textbf{comprehension}, where an unpredictable (or predictable) outcome is presented in the material (and controlled for frequency and event typicality). Using self-paced reading, participants show varying reading times near the target, which reflects processing ease. However, there are also studies on \textbf{production}, in which an outcome is not provided but elicited from participants through sentence completion (\autoref{other_studies}). They provide an additional set of contextual cues that future studies can examine.

\paragraph{Information Redundancy} Our introduction section presents another account in which highly predictable linguistic units from the local context can be difficult for humans to process, which we elaborate on here. When a conversation contains utterances that contain event sequences describing common activities (e.g., \textit{paying the cashier} when \textit{shopping}) \cite{kravtchenko2022informationally}, or are trivial from the context \cite{rees2026you}, these utterances become redundant to an extent that violates the Gricean expectation that utterances should be informative and relevant. Therefore, comprehenders may repair this dip in information utility by drawing pragmatic inferences to recover why a speaker would have made such an utterance. 

\begin{table*}[htbp]
\small
\begin{tabular}{lll|cc}
\toprule
 & \textbf{Cue for Informativity} & \textbf{Example} & \textbf{Compre.} & \textbf{Prod.}                 \\ \midrule

\citet{he2026just} & Adverbial ``Surprisingly'' & \textit{Surprisingly}, the house doesn't have ... & \checkmark &     
\\ \midrule
\citet{reksnes2024tell} & Speaker salience & \textit{At/ They are/ I am} at the train station... & & \checkmark    
\\
& Speaker style & \makecell[l]{Hi! What's up? \\ 
\textbf{Low-informativity speaker}: Hey, I just had breakfast. \\
\textbf{High-informativity speaker}: I'm at a golf course, \\and there's a celebrity here. 
}& & \checkmark \\  \midrule
\citet{rohde2022better}      & Reported thought or speech  & Rebecca \textit{thinks/ announces to me}... & & \checkmark                
\\
\textbf{}            & Likelihood of speech & \textit{when asked about it/ out of the blue}, Rebecca said... & & \checkmark
\\
\textbf{}            & Audience size & Rebecca said \textit{to me/ to everyone}... & & \checkmark \\

\bottomrule
\end{tabular}
\caption{A summary of additional informativity psycholinguistic studies that are not part of our evaluation.}
\label{other_studies}
\end{table*} 

\section{Expanded Dataset Construction}
\label{dataset_construction}

While the paper presents model results on stimuli from previous psycholinguistic studies, the sample size is relatively small ($N\leq25$ for each context type and condition), and it raises the concern whether models show consistent patterns in other examples with similar context setups. This section describes our dataset construction procedures. \S \ref{expanded_results} shows that model results in the paper are consistent with those using the expanded dataset. 

We expand the experimental stimuli from \citet{rohde2021s}, in which a subject used an instrument that was either conventional or unconventional for the task. We adapt \citet{jiang2021learning}, which contains actions that are common functions of human-made physical artifacts. This dataset allows us to ensure an instrument is highly conventional for a given task. 

In addition, we expand the stimuli in \citet{jiang2025expecting}, where a subject carried out a task at a conventional or unconventional location. We adapt \citet{mukuze2018vision}, which includes verb-location pairs that are rated for typicality by human annotators on a scale from 1 (extremely uncommon) to 7
(extremely common). We select pairs with a score of at least 5.5. 

For both datasets, we add a complement to each verb to provide a complete description of the task. (e.g., \textit{fly, airfield} $\rightarrow$ \textit{fly a plane, airfield}). We filter for tasks that are commonplace activities. We follow the stimulus design in previous studies and counterbalance the target so it appears in both informative and uninformative contexts exactly once. This procedure allows us to compare words with varying lengths and frequencies. We ensure that the unconventional target is unpredictable in context but physically possible to use for the task. In total, we construct 117 task-instrument pairs and 72 task-location pairs. We then apply them to their corresponding template to construct minimal pairs.

    \begin{quote}
        \small 
        \textbf{Unusual Protagonist}:
        My cousin Mary is a boring/ surprising person who always/never does things the way you’d expect. For instance, in order to \texttt{[Task]}, she was using a \texttt{[Tool]} yesterday in the afternoon.
        
        \textbf{Canonical}:
        My cousin Mary was \texttt{[Task]} yesterday in the afternoon. She was \texttt{[Task]} with a \texttt{[Tool]}, and she got really tired. 

        \textbf{Wh-cleft}:
        My cousin Mary was \texttt{[Task]} yesterday in the afternoon. What she was \texttt{[Task]} with was a \texttt{[Tool]}, and she got really tired. 

        \textbf{Unrealistic Scenario}:
        \texttt{[Protagonist]} is telling her friend what she did/ dreamt on Sunday. That day, she \texttt{[Task]} in \texttt{[Location]} before/ after she \texttt{[Secondary Task]}.
    \end{quote}

\section{An Alternative Theoretical Explanation}
\label{alternative_account}
\citet{futrell2012processing} provides an alternative account by examining the relationship between the allocation of processing resources and processing cost. When a context facilitates comprehenders to expect informative content, they preemptively allocate cognitive resources toward processing the upcoming material, resulting in processing ease for unpredictable outcomes. \citet{xu2025informativity} provide some evidence for this account:
More informative target nouns show weaker agreement attraction effects (measured by reading time of sentences varying in grammaticality and the number feature of distractors). The finding suggests that informative linguistic units are prioritized in working memory.

However, this account cannot directly explain why informativity effects often appear at the sentence wrap-up and not at the target.  When comprehenders encounter predictable outcomes, they may notice that resources were poorly allocated to uninformative content and resolve such conflicts by making pragmatic inferences \cite{kravtchenko2022informationally}, which may explain why informativity effects appear later than the target.

\section{General and Context-Dependent Plausibility}
\label{experiment1_supplementary}

We follow the procedures in \citet{kauf2024log} to examine whether the models we have chosen show general and context-dependent plausibility judgments as defined below:

\paragraph{General Plausibility} We assign the model a hit if it assigns higher probability to plausible sentences than to minimally different implausible sentences when no influencing context is present.

\paragraph{Context-Dependent Plausibility} We assign the model a hit if it assigns higher probability to implausible sentences with a licensing context than when no influencing context is present.

We evaluate model performance
through the likelihood they assign to either a critical word within the target sentence or the target
sentence as a whole. 

While all models achieve \textbf{full accuracy} for general plausibility in every setting, their performance for context-dependent plausibility varies by context type and model. We find that the unrealistic scenario context increases the plausibility of implausible sentences less often than the unusual protagonist condition does.     


\begin{table}[htbp]
  \centering
\resizebox{1\columnwidth}{!}{%
  \begin{tabular}{lcccc}
    \toprule
    & \multicolumn{2}{l}{\textbf{Unreal. Scenario}} & \multicolumn{2}{l}{\textbf{Unus. Protagonist}} \\
    \cmidrule(lr){2-3} \cmidrule(lr){4-5}
    \textbf{Model} & Tok & Sent & Tok & Sent \\
    \midrule
    Llama-3.1-8B Base & 0.458 & 0.375 & 0.769 & 0.769 \\
    Llama-3.1-8B Instruct   & 0.667 & 0.625 & 0.846 & 0.846 \\
    Qwen-2.5-7B Base  & 0.542 & 0.542 & 0.846 & 0.846 \\
    Qwen-2.5-7B Instruct    & 0.625 & 0.583 & 0.846 & 0.846 \\
    \bottomrule
  \end{tabular}
  }
  \caption{Context-Dependent Plausibility. \textbf{Tok} and \textbf{Sent} refer to token and sentence-level likelihood respectively.}
\end{table}

\section{Layerwise Surprisal Results}
\label{experiment1_layer}

We compute per-layer surprisal using \textit{logit lens} and \textit{tuned lens} on the same set of models in \S\ref{surprisal}.
We split model layers into 10 bins and calculate the standardized mean difference of target probability between each pair of conditions for three informativity effects. The results largely agree with those presented in \S\ref{surprisal}, that models fail to exhibit the reversal effect in all context types.  
\begin{figure}[htbp]
    \centering
    \includegraphics[width=1\linewidth]{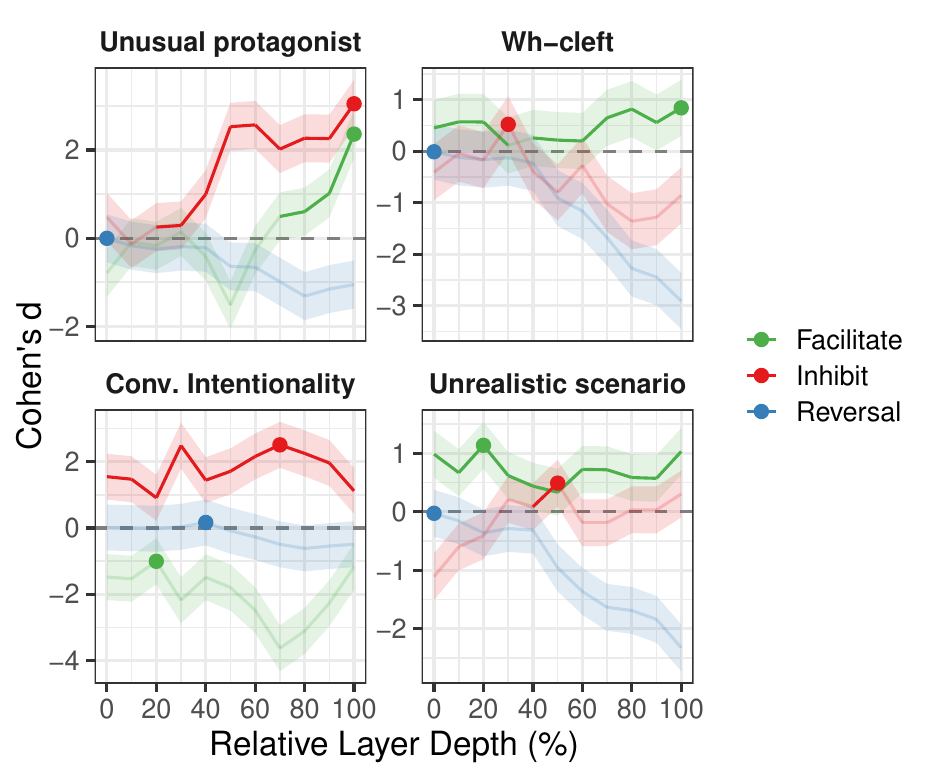}
    \caption{Standardized mean difference in lexical expectedness for the facilitation, inhibition, and reversal effects using layer-wise surprisal. A marker denotes the best-performing layer depth for each effect. Markers are transparent if their difference is $< 0.2$.}
\end{figure}

\newpage




\section{Sentence Completion Examples}
\label{completion_examples}
We provide completion examples for all contexts tested in \S \ref{sentence_completion} with Llama-3.1-8B in the direct measurement setting. The generation is ordered by probability. The examples in \autoref{completions} show that many completions under an informative context are still predictable and physically feasible to carry out the action described. The most probable generation uncovers the predictable target.
\begin{table}[h]
\centering
\small
\setlength{\tabcolsep}{4pt}
\renewcommand{\arraystretch}{1.15}
\begin{tabularx}{\linewidth}{@{}lX@{}}
\toprule
\textbf{Condition} & \textbf{Generation} \\
\midrule
\multicolumn{2}{@{}l@{}}{\makecell[l]{\textbf{Unusual Protagonist}\\
\textit{In order to dig a hole, she was using a ...}}} \\
\addlinespace[2pt]
Uninformative & \textcolor{black!100}{shovel}, \textcolor{black!64}{spade}, \textcolor{black!43}{spoon}, \textcolor{black!38}{fork}, \textcolor{black!35}{hammer}, \textcolor{black!35}{tool}, \textcolor{black!35}{hoe}, \textcolor{black!35}{broom}, \textcolor{black!35}{garden spade}, \textcolor{black!35}{pick}, \textcolor{black!35}{crowbar}, \textcolor{black!35}{metal utensil} \\
\addlinespace[2pt]
Informative & \textcolor{black!100}{shovel}, \textcolor{black!70}{spade}, \textcolor{black!59}{spoon}, \textcolor{black!59}{hoe}, \textcolor{black!53}{stick}, \textcolor{black!47}{fork}, \textcolor{black!47}{rock}, \textcolor{black!47}{rake}, \textcolor{black!35}{bowling ball}, \textcolor{black!35}{hammer drill}, \textcolor{black!35}{crowbar}, \textcolor{black!35}{paintbrush}, \textcolor{black!35}{toothpick}, \textcolor{black!35}{flower pot}, \textcolor{black!35}{screw}, \textcolor{black!35}{pick axe}, \textcolor{black!35}{snowplow}, \textcolor{black!35}{metal detector}, \textcolor{black!35}{needle}, \textcolor{black!35}{kitchen spoon}, \textcolor{black!35}{metal rod}, \textcolor{black!35}{dog}, \textcolor{black!35}{pick}, \textcolor{black!35}{metal rasp}, \textcolor{black!35}{pickax}, \textcolor{black!35}{garden rake}, \textcolor{black!35}{broom}, \textcolor{black!35}{hammer}, \textcolor{black!35}{straw}, \textcolor{black!35}{wheelbarrow} \\
\midrule
\multicolumn{2}{@{}l@{}}{\makecell[l]{\textbf{Unrealistic Scenario}\\
\textit{put some fresh meat and vegetables in the ...}}} \\
\addlinespace[2pt]
Uninformative & \textcolor{black!100}{refrigerator}, \textcolor{black!63}{fridge}, \textcolor{black!50}{car}, \textcolor{black!39}{freezer}, \textcolor{black!38}{kitchen}, \textcolor{black!37}{trunk}, \textcolor{black!35}{ice chest} \\
\addlinespace[2pt]
Informative & \textcolor{black!100}{refrigerator}, \textcolor{black!75}{fridge}, \textcolor{black!52}{freezer}, \textcolor{black!51}{car}, \textcolor{black!36}{shopping cart}, \textcolor{black!36}{storage}, \textcolor{black!36}{trunk}, \textcolor{black!36}{basket}, \textcolor{black!35}{cart}, \textcolor{black!35}{shopping basket}, \textcolor{black!35}{kitchen}, \textcolor{black!35}{washing machine} \\
\midrule
\multicolumn{2}{@{}l@{}}{\makecell[l]{\textbf{Wh-cleft}\\
\textit{What she was digging a hole with was a ...}}} \\
\addlinespace[2pt]
Uninformative & \textcolor{black!100}{shovel}, \textcolor{black!64}{spade}, \textcolor{black!39}{hoe}, \textcolor{black!35}{fork}, \textcolor{black!35}{stick}, \textcolor{black!35}{pick axe}, \textcolor{black!35}{digger}, \textcolor{black!35}{construction tool}, \textcolor{black!35}{pick}, \textcolor{black!35}{posthole digger}, \textcolor{black!35}{pickax}, \textcolor{black!35}{rock}, \textcolor{black!35}{wooden handle} \\
\addlinespace[2pt]
Informative & \textcolor{black!100}{shovel}, \textcolor{black!69}{spade}, \textcolor{black!50}{spoon}, \textcolor{black!50}{stick}, \textcolor{black!41}{teaspoon}, \textcolor{black!38}{trowel}, \textcolor{black!35}{wrench}, \textcolor{black!35}{garden shovel}, \textcolor{black!35}{pick}, \textcolor{black!35}{rake}, \textcolor{black!35}{plastic shovel}, \textcolor{black!35}{cane}, \textcolor{black!35}{machine}, \textcolor{black!35}{fork}, \textcolor{black!35}{garden digger}, \textcolor{black!35}{plastic trowel}, \textcolor{black!35}{knife}, \textcolor{black!35}{bar}, \textcolor{black!35}{garden tractor}, \textcolor{black!35}{metal shovel} \\
\bottomrule
\end{tabularx}
\caption{Completions by prompt type and condition. Word color indicates generation probability.}
\label{completions}
\end{table}

\newpage

\section{Full Results on Expanded Dataset}
\label{expanded_results}
To show that patterns found in model results are robust to instances beyond those from previous psycholinguistic studies \cite{rohde2021s,jiang2025expecting}, we replicate all experiments presented in the paper using the expanded dataset we constructed (with its procedure described in \autoref{dataset_construction}). We follow the same experimental setup as in the paper and run on Qwen-3-32B.

\begin{figure}[ht]
    \centering
    \includegraphics[width=1\linewidth]{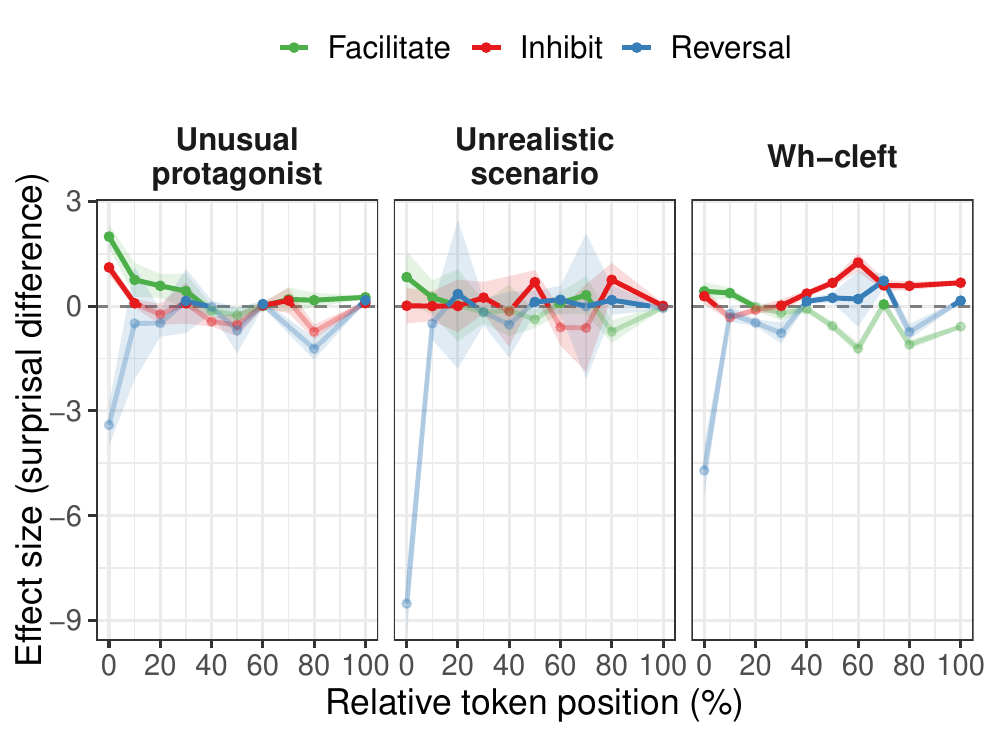}
    \caption{Informativity effect at words after the target
using standardized surprisal differences with 95\% confidence intervals.}
    \label{experiment1_wrapup}
\end{figure}

\paragraph{Experiment 1: Surprisal fails to align with informativity} Consistent with the results in the main paper, we find no clear evidence that surprisal exhibits all three effects at a token position across context types, whether at the target or at words after the target. (\autoref{experiment1_wrapup}). At the target, models exhibit strong negative reversal effects across context types, specifically in the unrealistic scenario context (similar to the pattern found in \autoref{wrapup}).   
\paragraph{Experiment 2: LLM expects informative continuations}
Consistent with the results in the main paper, we find that LLMs produce more variable and atypical completions under informativity expectations. (\autoref{experiment2b}). 
\begin{table}[htbp]
\centering
\resizebox{0.7\columnwidth}{!}{%
\begin{tabular}{lccc|c}
\toprule
& \multicolumn{3}{c}{\textbf{Var} ($\uparrow$)} & \multicolumn{1}{c}{} \\
\cmidrule(lr){2-4}
 & Tok & Comp & Rel & \multicolumn{1}{c}{\textbf{Typ} ($\downarrow$)} \\
\midrule
\textbf{Unusual Protagonist}  & \cellcolor{green!12}0.51 & \cellcolor{green!27}0.47 & \cellcolor{green!12}7.90  & \cellcolor{green!12}-0.13 \\
\textbf{Unrealistic Scenario} & \cellcolor{green!88}1.35 & \cellcolor{green!88}0.89 & \cellcolor{green!88}12.90 & \cellcolor{green!88}-0.41 \\
\textbf{Wh-cleft}             & \cellcolor{green!28}0.69 & \cellcolor{green!12}0.37 & \cellcolor{green!65}11.39 & \cellcolor{green!83}-0.39 \\
\bottomrule
\end{tabular}%
}
\caption{Standardized mean difference for the variability (\textbf{Var}) and typicality (\textbf{Typ}) of completions between informative and uninformative contexts ($c_1 - c_0$). Green cells indicate expected effect sizes, and red cells indicate unexpected ones. Results from Qwen3-32B with completion-based elicitation.}
\label{experiment2b}
\end{table}

\paragraph{Experiment 3: Post-hoc inferences align with informativity}
\begin{figure}[htbp]
    \centering
    \includegraphics[width=\linewidth]{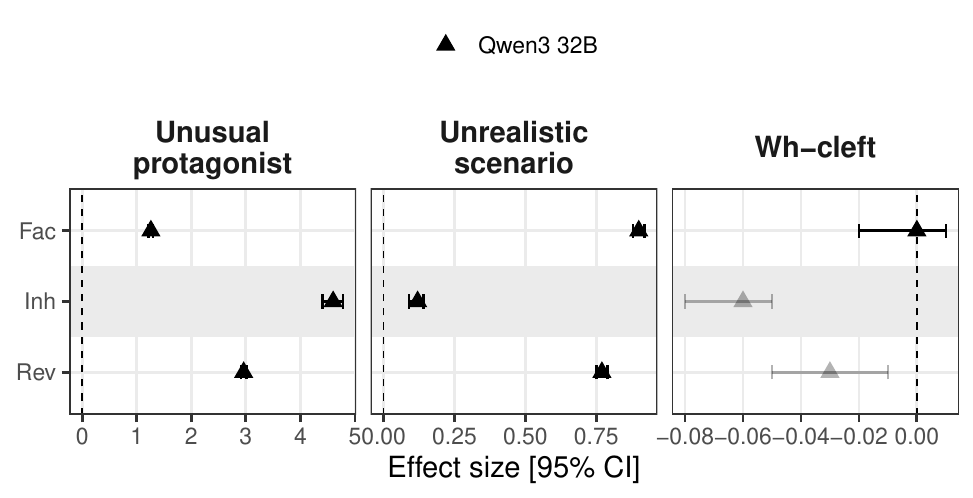}
    \caption{Standardized mean difference of verbalized expectedness for informativity effects with $95\%$ confidence interval. Positive effect sizes are expected.}
    \label{experiment3b}
\end{figure}
Consistent with the results in the main paper, we find that LLMs exhibit all three informativity effects in semantically explicit contexts (unusual protagonist and unrealistic scenario), and that these effects are stronger in the unusual protagonist context. However, we do not find the \cred{\textbf{inhibition}} or \cblue{\textbf{reversal}} effect in the wh-cleft context, and informativity effects have the largest variance in this context. The result provides further evidence that informativity effects are inconsistent with more implicit cues.

\newpage

\section{Prompts}

\label{verbalized_expectedness_prompts}
\paragraph{Typicality Pre-test}
To assess the typicality of a variety of target words, we use a reference model to list the five most likely words it would expect in a given event. These completions are instruments, as in the \textit{unusual protagonist} and \textit{wh-cleft} conditions, and locations, as in the \textit{unrealistic scenario} condition. 

\begin{figure}[h]
\centering
\resizebox{0.9\columnwidth}{!}{%
\begin{tcolorbox}[
  colback=brown!10,
  colframe=black,
  boxrule=0.5pt,
  sharp corners,
  width=\columnwidth
]
\ttfamily
\small
Complete the following sentence with five nouns that would best fit in the context:\newline

In order to dig a hole, Mary was using \_\_\_.
\end{tcolorbox}
}
\caption{An example prompt for the typicality pre-test where the target is an instrument.}
\end{figure}

\paragraph{Sentence Completion} We provide the metalinguistic prompt for the sentence completion task in \autoref{sent_completion_prompt}.

\begin{figure}[h]
\centering
\resizebox{0.9\columnwidth}{!}{%
\begin{tcolorbox}[
  colback=brown!10,
  colframe=black,
  boxrule=0.5pt,
  sharp corners,
  width=\columnwidth
]
\ttfamily
\small
Fill in the blank.\textbackslash{}n

In order to dig a hole, Mary was using \_\_\_.\textbackslash{}n

Answer:

\end{tcolorbox}
}
\caption{An example prompt for the sentence completion where the target is an instrument.}
\label{sent_completion_prompt}
\end{figure}

\paragraph{Verbalized Expectedness} We adapt prompts from \citet{kamath2026humans} for probabilistic inference and probe the following: (1) \textbf{Likelihood}: how likely the model considers the target as a continuation of the context; (2) \textbf{Coherence}: how consistent the target is with the context semantically; (3) \textbf{Felicity}: how appropriate the target is as a contribution to the discourse; (4) \textbf{Theory of Mind}: how well the target matches the speaker's intent in the context as inferred by the model. The system and user prompts are respectively \autoref{fig:verbalized-system-prompts} and \autoref{fig:verbalized-prompts}.

\begin{figure}[htbp]
\centering
\resizebox{0.9\columnwidth}{!}{%
\begin{tcolorbox}[
  colback=brown!10, colframe=black, boxrule=0.5pt, sharp corners,
  width=\columnwidth, left=4pt, right=4pt, top=4pt, bottom=4pt
]
\ttfamily\small
\slot{persona} \slot{task}
After any internal reasoning, reply with a single number between 0 and 100,
enclosed in \textless answer\textgreater{} tags.\newline
You can use the following descriptions of numerical ranges to help guide your
response:\newline
0: Completely \slot{neg}\newline
1--5: Almost completely \slot{neg}\newline
6--15: Highly \slot{neg}\newline
16--34: \slot{neg}\newline
35--49: Somewhat \slot{neg}\newline
50: \slot{mid}\newline
51--65: Somewhat \slot{pos}\newline
66--84: \slot{pos}\newline
85--94: Highly \slot{pos}\newline
95--99: Almost completely \slot{pos}\newline
100: Completely \slot{pos}
\end{tcolorbox}%
}%
\par\vspace*{10pt}\par
{\footnotesize
\setlength{\tabcolsep}{4pt}
\renewcommand{\arraystretch}{1.15}
\resizebox{\columnwidth}{!}{%
\begin{tabularx}{\columnwidth}{@{}
    >{\raggedright\arraybackslash}p{0.23\columnwidth}@{\hspace{5pt}}
    >{\scshape\raggedleft\arraybackslash}p{0.15\columnwidth}@{\hspace{6pt}}
    >{}X@{}}
\toprule
\textbf{Probe} & \multicolumn{2}{c}{\textbf{\rmfamily Slot fillers}} \\
\midrule
\textbf{Likelihood}
  & task    & You provide responses to questions about the expectedness of a
              continuation given some situation. \\
  & neg     & unexpected \\
  & pos     & expected \\
  & mid     & Totally even chance \\
\addlinespace[2pt]\midrule
\textbf{Coherence}
  & persona & You are a linguist with a strong understanding in discourse
              coherence. \\
  & task    & Your task is to judge whether a proposed continuation is a
              coherent ending for a given situation. \\
  & neg     & incoherent \\
  & pos     & coherent \\
  & mid     & Neither coherent nor incoherent \\
\addlinespace[2pt]\midrule
\textbf{Felicity}
  & persona & You are a very sociable person with a strong sense of pragmatic
              understanding. \\
  & task    & Your task is to judge whether a proposed continuation is an
              appropriate ending for the context described in a given
              situation. \\
  & neg     & inappropriate \\
  & pos     & appropriate \\
  & mid     & Neither appropriate nor inappropriate \\
\addlinespace[2pt]\midrule
\textbf{Theory of Mind}
  & persona & You are a very sociable person who is good at understanding a
              speaker's intention. \\
  & task    & Your task is to infer, using only the provided context, what the
              speaker most likely intends to say next, and then evaluate how
              well the actual continuation matches that expected message. \\
  & neg     & unexpected \\
  & pos     & expected \\
  & mid     & Totally even chance \\
\bottomrule
\end{tabularx}%
}%
}
\caption{System prompts for verbalized expectedness. All four conditions share
the template above and differ only in the marked \slot{slots}.}
\label{fig:verbalized-system-prompts}
\end{figure}

\begin{figure}[htbp]
\centering
\resizebox{0.9\columnwidth}{!}{%
\begin{tcolorbox}[
  colback=brown!10, colframe=black, boxrule=0.5pt, sharp corners,
  width=\columnwidth, left=4pt, right=4pt, top=4pt, bottom=4pt
]
\ttfamily\small Consider the following situation and continuation.\newline
Situation: \newline Continuation: \newline
\slot{query} Respond with a numerical value between 0 and 100,
where 0 indicates that \slot{low}, and 100 indicates that
\slot{high}. \slot{scope}
\end{tcolorbox}%
}%
\par\vspace*{10pt}\par
{\footnotesize
\setlength{\tabcolsep}{4pt}
\renewcommand{\arraystretch}{1.15}
\resizebox{\columnwidth}{!}{%
\begin{tabularx}{\columnwidth}{@{}
    >{\raggedright\arraybackslash}p{0.23\columnwidth}@{\hspace{5pt}}
    >{\scshape\raggedleft\arraybackslash}p{0.105\columnwidth}@{\hspace{6pt}}
    >{}X@{}}
\toprule
\textbf{Probe} & \multicolumn{2}{c}{\textbf{\rmfamily Slot fillers}} \\
\midrule
\textbf{Likelihood}
  & query & Given the situation, how much do you expect this continuation? \\
  & low   & you definitely did not expect the continuation \\
  & high  & you definitely expected the continuation \\
\addlinespace[2pt]\midrule
\textbf{Coherence}
  & query & How coherent is the continuation as an ending to the given situation? \\
  & low   & the continuation is completely incoherent \\
  & high  & the continuation is completely coherent \\
  & scope & You should only consider the coherence of the continuation in relation
            to the situation, but not the coherence of the situation itself. \\
\addlinespace[2pt]\midrule
\textbf{Felicity}
  & query & How appropriate is the continuation as an ending to the context
            described in the situation? \\
  & low   & the continuation is completely inappropriate \\
  & high  & the continuation is completely appropriate \\
  & scope & You should only judge whether the continuation is appropriate given
            the situation, not whether the situation itself is appropriate. \\
\addlinespace[2pt]\midrule
\textbf{Theory of Mind}
  & query & Using the context alone, predict what the speaker most likely intends
            to say next. Then judge how well the actual continuation matches that
            prediction. \\
  & low   & the continuation definitely does not match your expected intent \\
  & high  & the continuation definitely matches your expected intent \\
\bottomrule
\end{tabularx}%
}%
}
\caption{User prompts for verbalized expectedness. All four conditions share the
template above and differ only in the marked \slot{slots}.}
\label{fig:verbalized-prompts}
\end{figure}
\newpage 
\section{Model Inference}
\label{exp3_inference_detail}
We run inference on GPT-5 (\texttt{gpt-5-2025-08-07}) using the OpenAI API and Qwen3 235B (\texttt{Qwen3-235B-A22B-Instruct-2507-tput}) using the Together AI API. For both models, we made API calls using each respective provider’s Batch API functionality. Temperature values were set to model defaults. Reasoning effort is set to low, medium and high for GPT-5 and Qwen3 235B. We run inference on Qwen3-32B locally using vLLM with four NVIDIA A100 80GB. 

\section{Verbalized Expectedness Analysis}
\begin{figure}[htbp]
    \centering
    \includegraphics[width=1\linewidth]{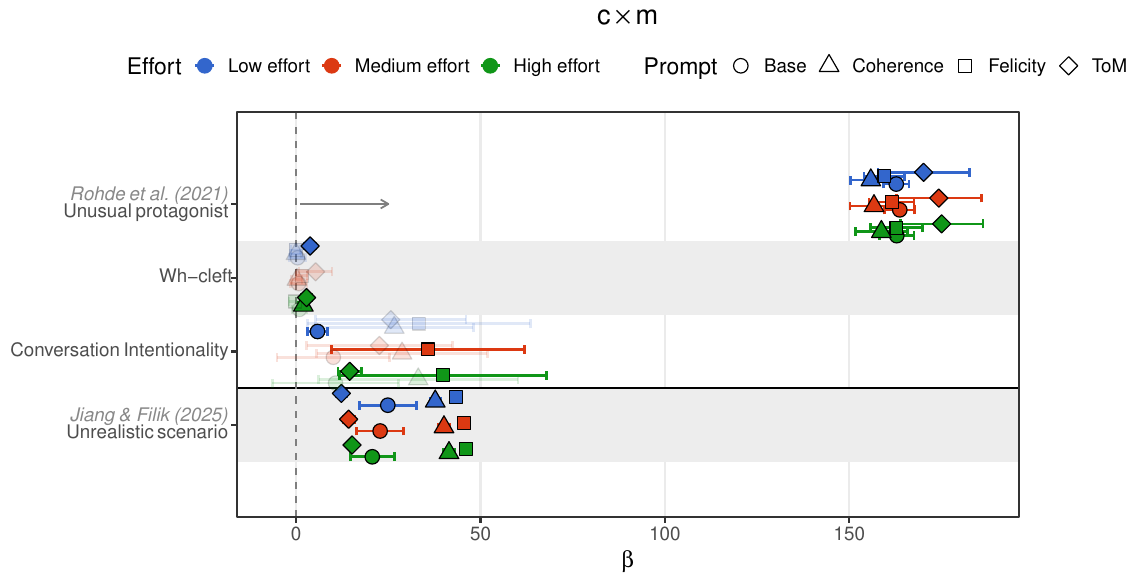}
    \caption{Interaction effects from linear mixed-effects models for verbalized expectedness with GPT-5 at varying reasoning effort (95\% CI). {\color{gray} Gray arrow} indicates the trend in human studies. $\beta$ is transparent if $p\geq0.05$ after FDR correction.}
    \label{experiment3_interaction}
\end{figure}

\begin{figure*}[htbp]
    \centering
    \includegraphics[width=0.8\linewidth]{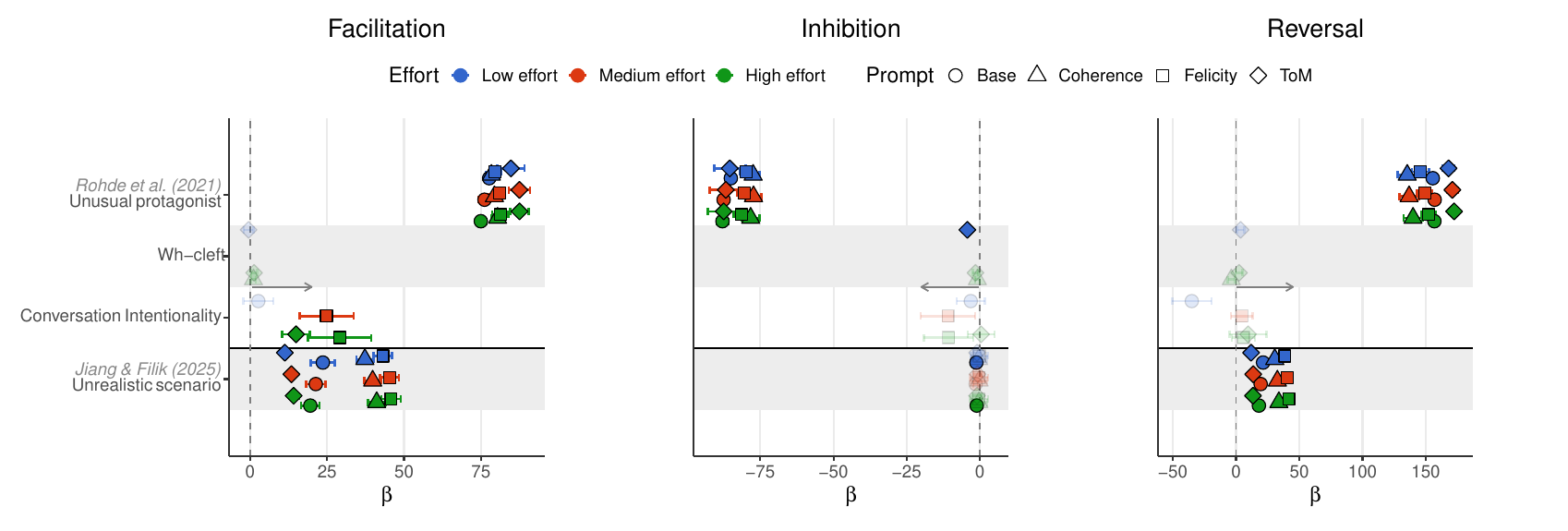}
    \caption{Conditional effects for verbalized expectedness with GPT-5 at varying reasoning effort (95\% CI). {\color{gray} Gray arrows} indicate the trends in previous human studies. $\beta$ is transparent if $p\geq0.05$ after FDR correction. See examples of conditional effect in \autoref{informativity_example}.}
    \label{experiment3_conditional}
\end{figure*}

In addition to effect sizes reported in the main paper, we conduct analysis of expectedness estimates in \S \ref{experiment3} with 
linear mixed-effects regression models, using the \verb|lme4| package in R. Fixed effects include $c \in \{c_0, c_1\}$ (coded as -0.5 and 0.5) and $m \in \{m_0, m_1\}$ (coded as -0.5 and 0.5). The models include by-item random intercepts and slopes. We test $c\times m$ interaction, where its significance is determined using Satterthwaite’s method of approximating degrees of freedom in the \verb|lmerTest| package. For all significant interactions, we then conduct post hoc analyses of simple effects to confirm informativity effects defined in \S \ref{task_datasets}. \autoref{experiment3_conditional} shows this result. We use the \verb|emmeans| package. The degrees of freedom in the post hoc tests were computed using the Satterthwaite approximation, with Tukey adjustment. We apply Benjamini-Hochberg false discovery rate correction to multiple comparisons in models and experimental conditions.

Congruent with human reading time results ($\hat{\beta} =-277.18$ and -0.07 respectively), the interaction effect is stronger in the unusual protagonist condition than in the unrealistic scenario condition (\autoref{experiment3_interaction}).




\begin{figure*}[t]
\begin{tcolorbox}[
  colback=gray!3,
  colframe=black,
  boxrule=0.5pt,
  sharp corners,
  width=\textwidth
]
\footnotesize

{\textbf{\normalsize What you'll see}}\\[0pt]
You'll be shown three things:
\begin{enumerate}[leftmargin=1.4em,itemsep=1pt,topsep=2pt]
  \item A \textbf{premise} --- a short piece of text.
  \item A \textbf{hypothesis} --- a likely continuation of that premise.
  \item A \textbf{summary} --- a short write-up of how one participant
        reasoned about whether the hypothesis is a good fit for the premise.
\end{enumerate}

\vspace{2pt}

\textbf{\normalsize Your task}\\[0pt]
Read the participant's \textbf{summary} and look for the five reasoning steps
described below. For each step:
\begin{itemize}[leftmargin=1.4em,itemsep=1pt,topsep=2pt]
  \item Decide whether that step appears in the summary, and give it a
        score using the scale provided.
  \item If the step is present, copy the exact part(s) of the summary
        that show it. If more than one part supports the step, copy all
        of them.
\end{itemize}
A step may or may not be present --- that's expected. Just score what you find.

\vspace{2pt}
\hrule
\vspace{2pt}

\textbf{Step 1 --- Did they notice a clue in the premise?}\\[0pt]
Look for signs that the participant picked up on something in the premise
that shaped what kind of continuation they expected. This could be:
\begin{itemize}[leftmargin=1.4em,itemsep=1pt,topsep=2pt]
  \item A clue about how informative the continuation should be --- for
        example, the personality of the main character, a sentence
        structure that emphasizes one part of the sentence, the
        intentions people express in conversation, or an unrealistic
        setting such as a dream.
  \item Any other clue the participant quotes directly from the premise.
\end{itemize}
\emph{Score:}
\begin{itemize}[leftmargin=1.4em,itemsep=1pt,topsep=2pt]
  \item[\textbf{0}] They don't mention or consider any clue from the premise.
  \item[\textbf{1}] They mention a clue that points toward an ordinary,
        predictable continuation.
  \item[\textbf{2}] They mention a clue that points toward an unusual,
        surprising continuation.
\end{itemize}

\vspace{1pt}
\textbf{Step 2 --- Did they judge how ordinary the hypothesis is on its own?}\\[0pt]
Look for a moment where the participant judges how expected or surprising
the hypothesis is by itself --- using plain common sense, before
factoring in the clues from Step 1. At this point they're only reacting
to how well the hypothesis fits its own immediate sentence, not the wider
context.\\[0pt]
\emph{Score:}
\begin{itemize}[leftmargin=1.4em,itemsep=1pt,topsep=2pt]
  \item[\textbf{0}] They don't comment on how ordinary or surprising the
        hypothesis is.
  \item[\textbf{1}] They judge the hypothesis to be ordinary or
        predictable on its own.
  \item[\textbf{2}] They judge the hypothesis to be unusual or
        unpredictable on its own.
\end{itemize}

\vspace{1pt}
\textbf{Step 3 --- Did they compare the hypothesis to alternatives?}\\[0pt]
Look for cases where the participant names one or more alternative
continuations that would also fit the premise, and compares them to the
original hypothesis to help make their judgment.\\[0pt]
\emph{Score:}
\begin{itemize}[leftmargin=1.4em,itemsep=1pt,topsep=2pt]
  \item[\textbf{0}] They don't mention any alternative.
  \item[\textbf{1}] They mention an alternative that is more ordinary
        than the hypothesis.
  \item[\textbf{2}] They mention an alternative that is more unusual
        than the hypothesis.
\end{itemize}

\vspace{1pt}
\textbf{Step 4 --- Did they decide the hypothesis fits the overall context?}\\[0pt]
Look for the point where the participant brings everything together ---
including the clues from Step 1 --- and judges whether the hypothesis
fits the overall theme.\\[0pt]
\emph{Score:}
\begin{itemize}[leftmargin=1.4em,itemsep=1pt,topsep=2pt]
  \item[\textbf{0}] They judge the hypothesis as not fitting the context overall.
  \item[\textbf{1}] They judge the hypothesis as fitting the context overall.
\end{itemize}

\vspace{1pt}
\textbf{Step 5 --- Did they explain away a surprising hypothesis?}\\[0pt]
Look for the participant offering an extra explanation for why a
surprising hypothesis could still make sense. They might, for example:
\begin{itemize}[leftmargin=1.4em,itemsep=1pt,topsep=2pt]
  \item guess the speaker's intent or personality,
  \item read the sentence non-literally,
  \item resolve a word with more than one meaning,
  \item lean on some other detail of the situation, or
  \item refer to a fictional or imaginary world.
\end{itemize}
The key is that they draw an additional inference not stated directly in
the premise in order to justify the hypothesis.\\[0pt]
\emph{Score:}
\begin{itemize}[leftmargin=1.4em,itemsep=1pt,topsep=2pt]
  \item[\textbf{0}] No such explanation is offered.
  \item[\textbf{1}] They offer at least one explanation for why the
        hypothesis does not fit.
  \item[\textbf{2}] They offer at least one explanation for why the
        hypothesis does fit.
\end{itemize}

\end{tcolorbox}
\caption{Annotation guideline for inference steps}
\label{annotation_guideline}
\end{figure*}

\section{Informativity Inference Steps}
\label{additional_inference_analyis}
We provide additional analyses on inference steps we define in \S \ref{inferences_informativity} using the excerpts of reasoning chains quoted below that are taken from the unusual protagonist context.

\paragraph{Anticipate informativity using contextual cues.} Before the model attends to the target word, it uses both the cue for informativity~(a) and other discourse cues~(b) to anticipate whether the continuation will be informative.
Failing to attend to the informativity cue or to identify its role in raising the expectation of informativity is the model's main failure. For instance, in the wh-cleft condition of \citet{rohde2021s}, the model fails to identify the cleft structure as an informativity cue. It does not attend to the cue or simply believes the cleft sounds unnatural (c).
\begin{quote}
    \small
    (a) Given that cousin Mary is seen as \hl{boring and predictable}
    \newline
    (b) \hl{``for instance''} suggests this is a typical example of her unexpected behavior
    \newline
    (c) The sentence ``what she was digging a hole with was a fork'' \hlr{comes off as unnatural} but could work for comedic effect.
\end{quote}

\paragraph{Identify the typicality of the target, then judge its felicity.} The model first identifies the typicality of the target word irrespective of the context using commonsense knowledge ~(d, ~e), then it judges the felicity of the target (i.e. whether it is an appropriate continuation to the discourse) respective of the context~(f). The model does so by updating its belief of the situation being described.
However, the model sometimes discounts the expectedness of the target because the cue raises informativity (g).
\begin{quote}
    \small
    (d) an unusual tool like a fork seems \hl{unexpected initially}
    \newline
    (e) to use something \hl{conventional}, like a shovel
    \newline
    (f) However, with her surprising nature, it actually becomes a \hl{plausible} choice.
    \newline
    (g) but given that she doesn't do things as expected, a shovel seems \hl{unlikely}.
    \newline
\end{quote}

\paragraph{Compare with alternatives to arrive at expectedness judgments.} To determine an exact expectedness score, the model usually first determines one or multiple numerical ranges using their description (e.g., \textit{6-15: Highly unexpected}. See \autoref{verbalized_expectedness_prompts}) and then narrows them down by reasoning over alternatives of the target word ~(h). The model proposes alternatives for both predictable and unpredictable target ~(i,~j) and compares them with the target for contextual coherence.
\begin{quote}
    \small
    (h) I might rate it between 78-88... But if I focus on ``a fork'' specifically, I might go lower, around 78, since \hl{a spoon} is a common trope for digging.
    (i) I can anticipate something unusual, like a \hl{spoon} or \hl{toy shovel}.
    \newline
    (j) Normally, tools like \hl{a shovel} or \hl{spade} come to mind for that task.
    \newline
    
\end{quote}

\paragraph{Accommodate for unpredictable target.} The model sometimes comes up with an explanation as to why the unpredictable target is felicitous with respect to the context. This inference step is analogous to atypicality accommodation in \citet{kravtchenko2022informationally}: comprehenders resolve why a speaker would have chosen to mention the situation in the first place. It may interpret non-literally (k, p), resolve lexical ambiguity of the target (l), use previous context to rationalize the behavior ~(n, ~p), or make references to fictional worlds ~(o):

\begin{quote}
    \small
    (k) using a fork here \hl{adds humor} due to its contradiction.
    \newline
    (l) A \hl{garden fork} might actually be somewhat normal for soil loosening
    \newline
    (m) It seems low but not impossible; \hl{maybe a child or someone limited in resources} might use it.
    
\end{quote}

\begin{quote}
    \small
\textbf{Unrealistic Scenario}: \textit{He came to his office by bike, ..., grabbed a cup of coffee, ... , and then wrote down some important tasks on a towel ...}
    \newline
    \newline
    (n) \hl{The mention of "coffee"} could relate to items like paper towels or napkins.
    
    (o) A towel seems odd in an office context, although \hl{it might be a Hitchhiker's Guide reference}, or maybe if someone just biked to work and had a gym towel.
    
    (p) While dreams often include bars, flu vaccinations don’t typically happen there. However, \hl{the pun on “shots” might influence the expectation}.
\end{quote}

\section{Inference Step Annotations}
\label{annotation}
\begin{figure*}[htbp]
    \centering
    \includegraphics[width=0.69\linewidth,height=0.85\textheight,keepaspectratio]{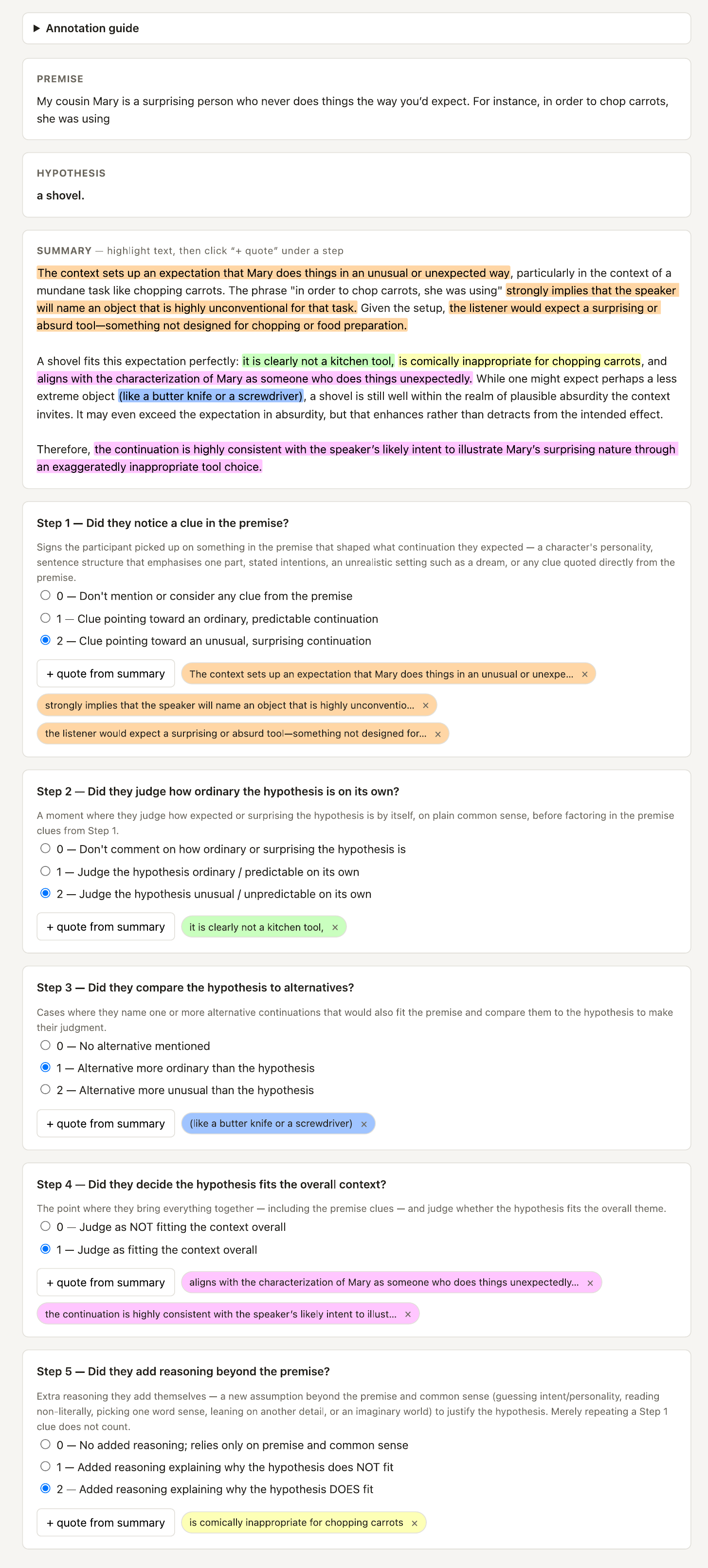}
    \caption{Annotation interface for inference steps}
    \label{annotation_ui}
\end{figure*}

\paragraph{Human Annotations} We recruit four human annotators who are native or
near-native adult English speakers. They are student volunteers with at least a master’s
degree in computer science, computational linguistics, or linguistics. They are first provided with the annotation guideline that provide definitions of both annotation task and inference steps (\autoref{annotation_guideline}). The annotation is completed on a web interface (\autoref{annotation_ui}). A working example is provided at the start for annotators to be familiar with the task. Then, participants are asked to identify whether an inference step is present in the summary passage. If a step is present, they need to highlight the exact part(s) of the passage that support an inference step. The average inter-annotator agreement between human and LLM is $\kappa = 0.79$, while the average pairwise agreement between any two human annotators is $\kappa = 0.78$.

\paragraph{LLM Annotations} The annotation procedures for Claude Opus 4.8 is similar, where the annotation guideline is provided as the system prompt. For each inference step, we enforce the output to contain (1) a score, (2) extracts from summary to support that inference step and additionally (3) an explanation as to why chosen extracts justify the score. We run inference on Claude Opus 4.8 using Anthropic API with default parameters.

\definecolor{sechdr}{gray}{0.96}
\newcommand{\secrow}[1]{%
  \multicolumn{5}{l}{\cellcolor{sechdr}\textbf{#1}}}
\newcommand{\ind}{\hspace{1em}}

\begin{table}[htbp]
\centering
\small
\resizebox{1\columnwidth}{!}{%
\begin{tabular}{lcccc}
\toprule
 & \makecell[c]{\textbf{Unus.}\\\textbf{Prot.}}
 & \makecell[c]{\textbf{Unre.}\\\textbf{Scen.}}
 & \makecell[c]{\textbf{Wh-}\\\textbf{Cleft}}
 & \makecell[c]{\textbf{Conv.}\\\textbf{Intent}} \\
\midrule
\secrow{Cue Detection} \\
\addlinespace[3pt]
\ind No cue detected                        & 0.05 & 0.09 & 0.48 & 0.10 \\
\ind Cue found \textbf{expected}            & 0.01 & 0.10 & 0.51 & \textbf{0.46} \\
\ind Cue found \textbf{surprising} $\checkmark$ & \textbf{0.94} & \textbf{0.80} & 0.01 & 0.44 \\
\addlinespace[4pt]
\secrow{Target Typicality} \\
\addlinespace[3pt]
\ind No typicality judgment                 & 0.04 & 0.05 & 0.01 & 0.04 \\
\ind Target judged typical                  & 0.03 & 0.01 & 0.00 & 0.19 \\
\ind Target judged atypical $\checkmark$    & \textbf{0.93} & \textbf{0.95} & \textbf{0.99} & \textbf{0.77} \\
\addlinespace[4pt]
\secrow{Target Alternatives} \\
\addlinespace[3pt]
\ind No alternative provided                & \textbf{0.52} & 0.36 & 0.15 & \textbf{0.71} \\
\ind \textbf{Typical} alternative           & 0.33 & \textbf{0.64} & \textbf{0.85} & 0.27 \\
\ind \textbf{Atypical} alternative          & 0.14 & 0.00 & 0.00 & 0.02 \\
\addlinespace[4pt]
\secrow{Target Fit} \\
\addlinespace[3pt]
\ind Doesn't fit context                    & 0.08 & \textbf{0.57} & \textbf{0.88} & 0.31 \\
\ind Fits context $\checkmark$              & \textbf{0.92} & 0.43 & 0.12 & \textbf{0.69} \\
\addlinespace[4pt]
\secrow{Accommodate} \\
\addlinespace[3pt]
\ind No accommodation                       & \textbf{0.54} & \textbf{0.45} & \textbf{0.40} & 0.23 \\
\ind Accom. target doesn't fit              & 0.06 & 0.25 & 0.27 & 0.17 \\
\ind Accom. target fits                     & 0.40 & 0.30 & 0.33 & \textbf{0.60} \\
\bottomrule
\end{tabular}
}
\caption{The distribution of inference steps for each context type. $\checkmark$ indicates the correct inference for that step.}
\label{step_distribution}
\end{table}

\paragraph{Data Consent and Ethics Consideration} Annotators provide consent and are informed about how their data would be used before participating in the study. They are told that their annotations can be deleted upon request. The study took approximately 60
minutes on average. We completed the ethical evaluation checklist provided by the university ethical review board, and the human study protocol raised no concerns.

\begin{table*}[htbp]
\footnotesize
\begin{tabularx}{\textwidth}{
l 
>{\raggedright\arraybackslash}p{2cm}   
>{\raggedright\arraybackslash}p{3cm} 
>{\raggedright\arraybackslash}X        
>{\raggedright\arraybackslash}X        
>{\raggedright\arraybackslash}p{2cm}   
}
\toprule
& \textbf{Cue} & \textbf{Premise Control} & \textbf{Premise} $c_0$ & \textbf{Premise} $c_1$ & \textbf{Continuation}\\ 
\midrule 
\citet{rohde2021s} 
& Unusual protagonist 
& Mary is my cousin. In order to dig a hole, she was using 
& My cousin Mary is a \textit{boring} person who always does things the way you’d expect. For instance, in order to dig a hole, she was using   
& My cousin Mary is a \textit{surprising} person who never does things the way you’d expect. For instance, in order to dig a hole, she was using 
&  \textit{a fork/ shovel}. \\ \cmidrule{2-6}

& Wh-cleft 
& My cousin Mary was digging a hole. She used
& My cousin Mary was digging a hole. She was digging the hole with 
& My cousin Mary was digging a hole. \textit{What she was} digging the hole with \textit{was} ... 
& \textit{a fork/ shovel}.\\ \cmidrule{2-6}

& Intentionality in conversation 
& 
Joe runs into Amanda and says it’s been a long time since they last saw each other. Amanda calls him a blast from the past, and asks how he’s been. Joe says not much but he thought of her earlier that day while he was at the mall. Amanda jokingly asks if he was buying her a present, and Joe jokes back that he wasn’t, but explains that he came across socks in her favorite shade of purple and wanted to tell her about it. The price for the socks was  

& \multicolumn{2}{l}{
\parbox[t]{4.3cm}{
\textbf{Joe}: hey, long time no see! 

\textbf{Amanda}: wow, a blast from the past! 

\textbf{Amanda}: how goes? 

\textbf{Joe}: not much happening 

\textbf{Joe}: but I thought of you today

\textbf{Joe}: I was at the mall

\textbf{Amanda}: :) buying a present for me?! 

\textbf{Joe}: :) no, but they had socks in your favorite shade of Mahogany 

\textbf{Joe}: I wanted to tell you what I saw – the price for the socks was actually
}
} 
& \textit{{\$150}/ {\$2}}.
\\ 
\midrule 
\citet{jiang2025expecting} 
& Unrealistic scenario 
& Mary met her friend. That day, she drove to the nearest grocery store with her husband, bought some fresh meat and vegetables, and then put them in
& Mary is telling her friend what she \textit{did} on Sunday. \textit{That day}, she drove to the nearest grocery store with her husband, bought some fresh meat and vegetables, and then put them in
& Mary is telling her friend what she \textit{dreamt} on Sunday. \textit{In her dream}, she drove to the nearest grocery store with her husband, bought some fresh meat and vegetables, and then put them in
& \textit{{the wardrobe}/ {refrigerator}.} \\
\bottomrule
\end{tabularx}
\caption{Full experimental conditions and stimuli for each informativity cue, including premise controls.}
\label{data_full}
\end{table*}

\end{document}